\documentclass[acmlarge]{acmart}
\usepackage{multirow}
\usepackage{threeparttable}
\usepackage{color}
\usepackage{soul}
\usepackage{wrapfig}
\usepackage{caption}
\usepackage{subcaption}
\usepackage{booktabs}
\usepackage{soul}
\usepackage{enumitem}
\usepackage{amsmath,amsfonts}
\usepackage{algorithmic}
\usepackage{graphicx}
\usepackage{textcomp}
\usepackage{xcolor}
\usepackage{makecell}
\usepackage{wrapfig}
\usepackage{xcolor}  
\newcommand{\name}{LanHAR}
\usepackage{color}

\AtBeginDocument{%
  }

\setcopyright{cc}
\setcctype{by}
\acmJournal{IMWUT}
\acmYear{2025} \acmVolume{9} \acmNumber{4} \acmArticle{230} \acmMonth{12} \acmPrice{}\acmDOI{10.1145/3770652}

\begin{document}

\title{Large Language Model-Guided Semantic Alignment for Human Activity Recognition}

\author{Hua Yan}
\affiliation{\institution{Lehigh University}
\country{USA}
}
\email{huy222@lehigh.edu}

\author{Heng Tan}
\affiliation{\institution{Lehigh University}
\country{USA}
}
\email{het221@lehigh.edu}

\author{Yi Ding}
\affiliation{\institution{University of Texas at Dallas}
\country{USA}
}
\email{yi.ding@utdallas.edu}

\author{Pengfei Zhou}
\affiliation{\institution{University of Pittsburgh}
\country{USA}
}
\email{pengfeizhou@pitt.edu}

\author{Vinod Namboodiri}
\affiliation{\institution{Lehigh University}
\country{USA}
}
\email{vin423@lehigh.edu}

\author{Yu Yang}
\affiliation{\institution{Lehigh University}
\country{USA}
}
\email{yuyang@lehigh.edu}

\begin{abstract}
Human Activity Recognition (HAR) using Inertial Measurement Unit (IMU) sensors is critical for applications in healthcare, safety, and industrial production. However, variations in activity patterns, device types, and sensor placements create distribution gaps across datasets, reducing the performance of HAR models. To address this, we propose \name, a novel system that leverages Large Language Models (LLMs) to generate semantic interpretations of sensor readings and activity labels for cross-dataset HAR. This approach not only mitigates cross-dataset heterogeneity but also enhances the recognition of new activities.
\name\ employs an iterative re-generation method to produce high-quality semantic interpretations with LLMs and a two-stage training framework that bridges the semantic interpretations of sensor readings and activity labels. This ultimately leads to a lightweight sensor encoder suitable for mobile deployment, enabling any sensor reading to be mapped into the semantic interpretation space.
Experiments on five public datasets demonstrate that our approach significantly outperforms state-of-the-art methods in both cross-dataset HAR and new activity recognition. The source code is publicly available at \url{https://github.com/DASHLab/LanHAR}.
\end{abstract}



\maketitle
\section{Introduction}
Human Activity Recognition (HAR) based on data collected from Inertial Measurement Unit (IMU) sensors on mobile platforms such as smartphones and wearable devices is one of the key problems in mobile computing due to its important applications in fields such as healthcare~\cite{subasi2018iot, subasi2020human}, safety~\cite{sun2022real, liagkou2022security,yan2023identifying}, and industrial production~\cite{niemann2021context, zheng2018comparison,yan2024robust}.
However, variations in activity patterns, device types, and sensor placements across different individuals result in significant distributional differences between datasets, even when capturing the same activity~\cite{xu2023practically, liu2022pred}.
Consequently, the performance of HAR models deteriorates considerably in cross-dataset human activity recognition scenarios~\cite{xu2023practically}.
Therefore, developing HAR models capable of generalizing across different datasets remains a critical research problem that needs solutions.

Existing work on cross-dataset HAR can be broadly categorized into two methods: domain adaptation and data augmentation. 
Domain adaptation methods~\cite{chang2020systematic,qin2019cross,khan2018scaling,zhou2020xhar} aim to bridge the distribution gap between source and target datasets to enhance model performance in training. 
Data augmentation methods~\cite{um2017data, xu2023practically, qian2022makes,saeed2019multi} seek to increase the diversity of training data to improve the model’s generalization ability.
While both methods are valuable, they still suffer from two limitations:
(i) Most existing work overlooks the physical meanings of activities.
This can exacerbate the impact of classification errors on downstream applications. 
For instance, misclassifying a walking activity as running may have a relatively minor effect compared to misclassifying it as a significantly different activity such as biking (see quantitative results in Evaluation).
(ii) Moreover, cross-dataset scenarios often introduce new activities that these methods struggle to handle, as these new activities are typically missing from the training phase.

\setlength{\columnsep}{6pt}
\begin{wrapfigure}{r}{0.35\linewidth}
    \centering
    \includegraphics[width=0.9\linewidth, keepaspectratio=true]{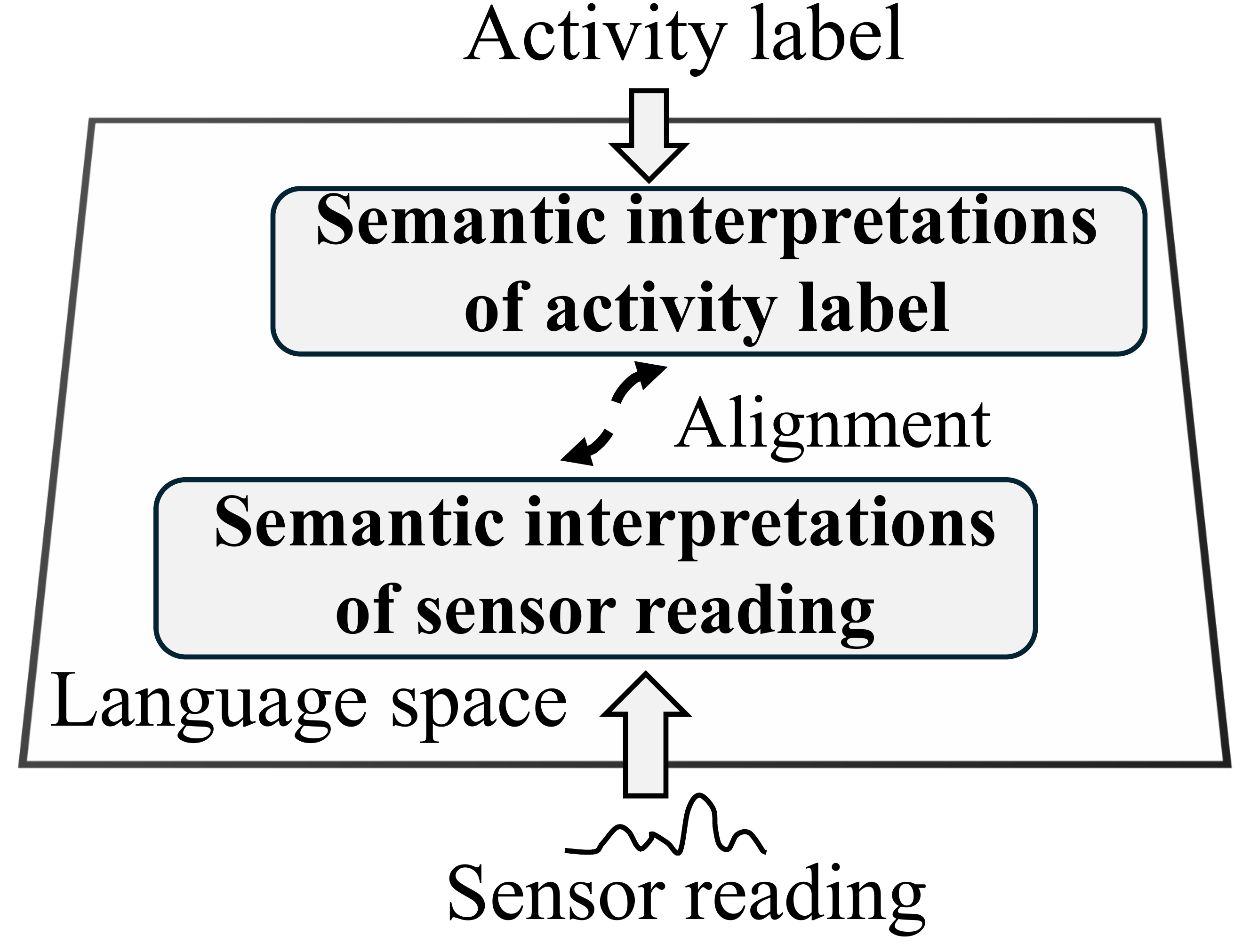}
    \vspace{-10pt}
    \captionsetup{font={normalsize}}
    \caption{Semantic interpretations for HAR}
    \label{fig:intro}
    \vspace{-10pt}
\end{wrapfigure}
We argue that semantic interpretations of sensor readings and activity labels can play a crucial role in addressing the aforementioned limitations as shown in Fig.~\ref{fig:intro}.
(i) For cross-dataset heterogeneity, interpreting sensor readings based on its physical meanings can help identify patterns associated with specific activities (e.g., jogging often exhibits regular periodicity).
The underlying intuition is that, despite the variability across datasets, activities may still reveal common underlying patterns. 
By identifying these patterns, we can map the heterogeneous data into a common space, thereby mitigating the impact of dataset variability.
(ii) When dealing with new activities, semantic interpretations of activity labels can enhance their meaning by generating detailed descriptions. 
Combining semantic interpretations of both sensor readings and activity labels creates the potential to map new data to new activity labels.
However, manual semantic interpretation is resource-intensive. The recent success of large language models (LLMs) offers a promising solution, enabling the generation of semantic interpretations from both sensor readings and activity labels through natural language~\cite{jiang2024motiongpt, zhang2023unleashing}. Research has shown that LLMs have the ability to perceive aspects of the physical world~\cite{xu2024penetrative}, providing an opportunity to leverage these capabilities to bridge the gap in cross-dataset HAR through automated semantic interpretation.

Despite the promise of semantic interpretations using LLMs, several challenges remain to be addressed:
(i) Generation Challenge: The semantic interpretations generated by LLMs can suffer from issues like hallucinations and randomness. Ensuring consistent and high-quality semantic interpretations remains a significant challenge.
(ii) Alignment Challenge: Once LLMs generate semantic interpretations of sensor readings and labels, the next step is to align them for human activity recognition, i.e., aligning sensor readings and labels based on their interpretations.
While language models can be used to align these interpretations, general language models lack specific knowledge of IMU data and human activity recognition, limiting their ability to accurately understand and reason about activities.
(iii) Deployment Challenge: 
Frequent access to cloud-based LLMs or deploying them locally on mobile devices is impractical due to high latency and resource constraints. 
A lightweight framework is needed to effectively leverage LLMs for on-device HAR.

To address these challenges, we design \name\, a novel system that leverages LLMs to generate semantic interpretations of sensor readings and activity labels, aiming to improve cross-dataset human activity recognition.
\name\ consists of three key components.  
(i) Generation: we carefully design prompts to guide LLMs in generating semantic interpretations and develop an iterative re-generation method to ensure high-quality semantic outputs.
(ii) Alignment: we design a semantic interpretation alignment module to enhance understanding of semantic interpretations based on a text encoder to encode interpretations and two contrastive learning tasks to improve the alignment between activities and sensor readings.
(iii) Deployment: We introduce a lightweight, two-stage training and inference framework. First, we train the semantic interpretation alignment module, then design and train a sensor reading encoder to map sensor data to the semantic interpretation space. For inference on mobile devices, we use the trained encoder to obtain sensor reading encodings and measure their similarity to activity label semantics to determine the HAR results.

In particular, our main contributions are as follows.
\begin{itemize}
\item 
We explore the use of LLM-generated semantic representations to improve cross-dataset human activity recognition. Our method maps both sensor readings and activity labels into a shared semantic space using natural language descriptions. This alignment reduces dataset heterogeneity and enables the recognition of activities that are not present in the training data.

\item
We design \name, a novel system that leverages LLMs to generate semantic interpretations of sensor readings and activity labels, addressing the challenge of cross-dataset human activity recognition. The system features a semantic interpretation generation process with an iterative re-generation method to ensure high-quality outputs, alongside a two-stage training framework that transfers the capabilities of large language models into lightweight, privacy-preserving models suitable for sustainable deployment on resource-constrained edge devices.

\item  We evaluate our system based on five public datasets.
The experimental results demonstrate that it outperforms other state-of-the-art methods in cross-dataset activity recognition, achieving an improvement of 7.35\% in accuracy and 13.16\% in F1 score. Additionally, for new activity recognition, it shows further enhancement, with improvements of 43.67\% in accuracy.
\end{itemize}

\section{Related work}
\subsection{Human activity recognition}
Human Activity Recognition (HAR) leverages data from various sensors such as accelerometers and gyroscopes to detect and classify human activities~\cite{yao2017deepsense,yang2015deep}.
To mitigate the issue of data heterogeneity, existing works can be classified into three categories, including semi-supervised learning methods, domain adaptation methods, and data augmentation methods.

In HAR, collecting large-scale labeled datasets is particularly challenging due to the high cost of manual annotation, while abundant unlabeled sensor data are readily available. To address this imbalance, semi-supervised methods~\cite{xu2021limu, saeed2019multi,zhang2024unimtsunifiedpretrainingmotion, miao2024goat, presotto2023combining} have been developed to maximize the use of unlabeled data by learning valuable feature representations without requiring explicit labels. These approaches enable models to adapt more efficiently to downstream tasks with only a minimal amount of labeled data, which is crucial for improving generalization across diverse datasets.
For example, UniMTS~\cite{zhang2024unimtsunifiedpretrainingmotion} is a unified pre-training framework designed for motion time-series data, with zero-shot and few-shot generalization across diverse human activity recognition tasks.
LIMU-BERT~\cite{xu2021limu} leverages masked language modeling and self-supervised learning to effectively utilize unlabeled IMU sensor data.
Despite their capabilities, these models still depend on a certain amount of labeled data to train classifiers, which may struggle when applied to target datasets that lack any labeled data.
GOAT~\cite{miao2024goat} uses pre-training with textual attributes to improve activity recognition across multiple datasets, supporting cross-dataset generalization.

The core idea of domain adaptation methods~\cite{chang2020systematic,qin2019cross,khan2018scaling,zhou2020xhar} involves taking a classifier that has been pre-trained on a source domain and using it with an unlabeled dataset from a target domain, then adjusting the source model's weights to enhance its performance in the target domain.
For example, SDMix~\cite{lu2022semantic} aims to mix features from the source and target domains in the feature space while emphasizing the preservation of semantic discrimination, in order to reduce the discrepancy between different domains.
UDAHAR~\cite{chang2020systematic} systematically studies the effectiveness of various unsupervised domain adaptation methods for human activity recognition tasks. It benchmarks multiple UDA techniques, such as feature alignment and adversarial adaptation, providing a comprehensive evaluation of their robustness under different domain shifts.
While domain adaptation methods mitigate domain shifts, they often struggle with large domain discrepancies, depend heavily on abundant source data

Data augmentation~\cite{um2017data, xu2023practically, qian2022makes,saeed2019multi,wang2022sensor,tang2021selfhar} in human activity recognition aims to increase the diversity and quality of training data, thereby enhancing the model's generalization ability and accuracy. 
For example, UniHAR~\cite{xu2023practically} designs physical-informed augmentation methods to improve the cross-dataset HAR performance.

However, All the methods are valuable, they still suffer from two limitations: (i) Most existing work overlooks the physical meanings of activities.
This can exacerbate the impact of classification errors on downstream applications. For instance, misclassifying a walking activity as running may have a relatively minor effect compared to misclassifying it as a significantly different activity such as biking. (ii) Moreover, cross-dataset scenarios often introduce new activities that these methods struggle to handle, as these new activities are typically missing from the training phase. Our potential idea to solve these limitations is to use and map semantic interpretations of sensor readings and activity labels in a shared semantic space to mitigate cross-dataset heterogeneity and generate new activity.

Recognizing new activities has been a longstanding challenge in HAR, particularly for machine learning-based
methods, which are unable to identify activities that were not present during the training phase. The most relevant method~\cite{cheng2013nuactiv} attempts to address the challenge of recognizing new activities by using pre-defined attributes, such as actions of specific body parts. The method trains two mappings: one from sensor readings to predefined attributes, and another from predefined attributes to activity labels. However, the quality of these predefined attributes significantly influences the model performance. Additionally, this work is not well-suited for cross-dataset new activities, as the new activities may differ significantly from the original dataset.

\subsection{LLM for human activity recognition}
Recently, there has been a surge of interest in leveraging the capabilities of LLMs~\cite{naveed2025comprehensive, jin2023time, tan2025llm} to enhance mobile computing applications~\cite{lee2024mobilegpt, wen2024autodroid}.
Penetrative AI~\cite{xu2024penetrative} explores the potential of leveraging large language models to comprehend and interact with the physical world. 
Kim et al.~\cite{kim2024health} harness the power of large language models to improve health prediction using data collected from wearable sensors.
IMUGPT2.0~\cite{leng2024imugpt} utilizes LLMs for virtual IMU data generation since IMU-labeled data is challenging to collect. It incorporate motion filtering and diversity metrics to enhance the efficiency and practicality of synthetic data generation for HAR. 
MotionGPT~\cite{jiang2023motiongpt} fuses language data with large-scale motion models, and motion-language pre-training that can enhance the performance of motion-related tasks becomes feasible.
HARGPT~\cite{ji2024hargpt} leverages LLMs as zero-shot human activity recognizers by framing sensor-based activity recognition as a natural language processing task.
ContextGPT~\cite{arrotta2024contextgpt} uses LLMs to extract commonsense knowledge about the environmental contexts of human activities, enhancing context-aware HAR with limited supervision.

Recent works exploring the use of large language models for human activity recognition have demonstrated the potential of LLMs to enhance HAR in various ways and have inspired our approach to leverage LLMs for generating semantic interpretations of IMU data. 
Nevertheless, while many studies explore LLMs in the context of HAR, a few directly address HAR tasks, and even those that still face challenges in enabling efficient deployment on resource-constrained mobile devices.

\section{Motivation}
\label{sec:motivation}
\subsection{LLMs possess the ability to perceive the physical world}
\begin{figure}[t] \centering
        \vspace{-10pt}
        \includegraphics[width=0.54\linewidth, keepaspectratio=true]{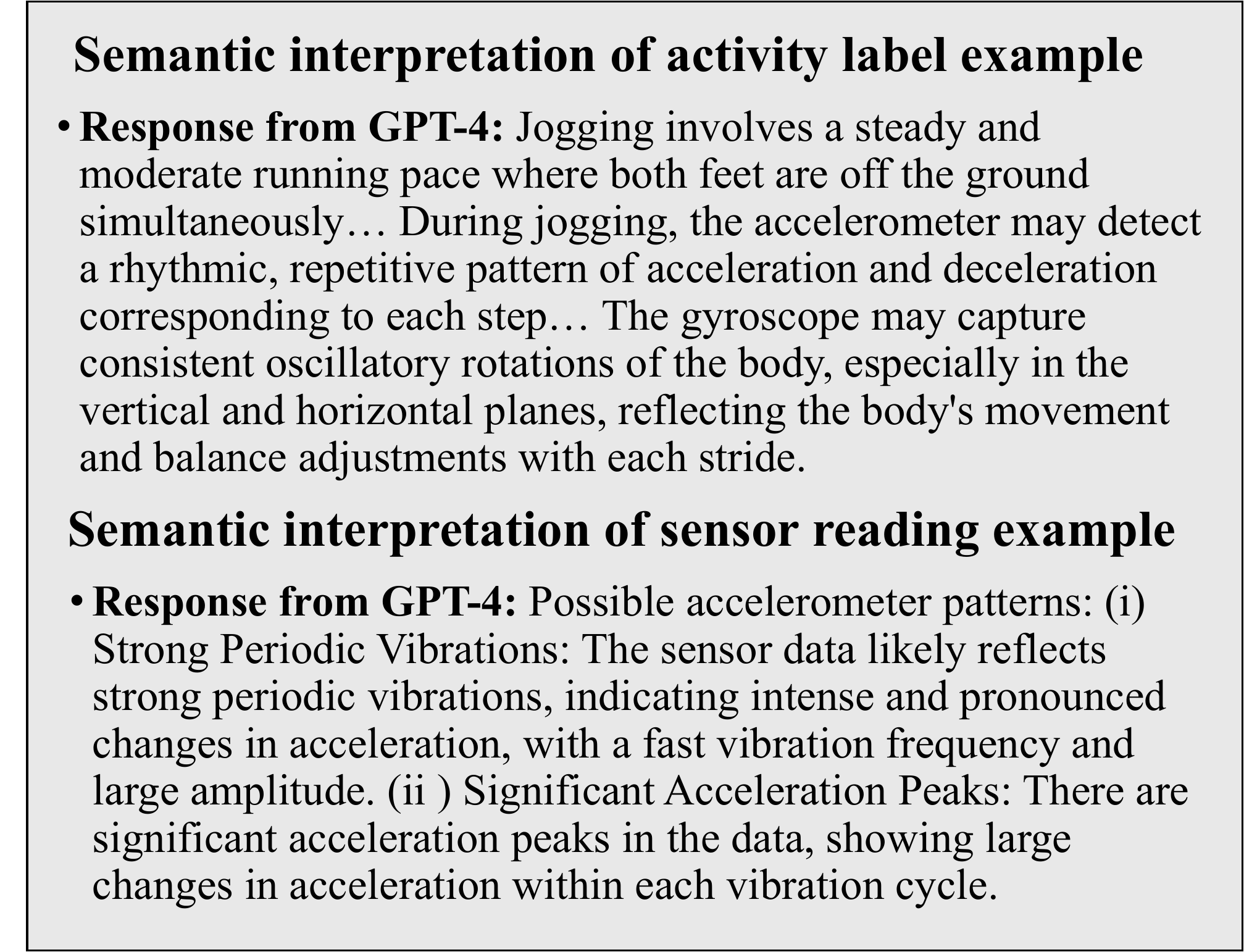}
        \captionsetup{font={normalsize}}
        \caption{Example of semantic interpretations of sensor readings and activity labels}
        \label{fig:example}
        \vspace{-10pt}
\end{figure}

\begin{figure*}[t] \centering
    \begin{subfigure}{0.3\linewidth} \centering
        \includegraphics[width=\linewidth, keepaspectratio=true]{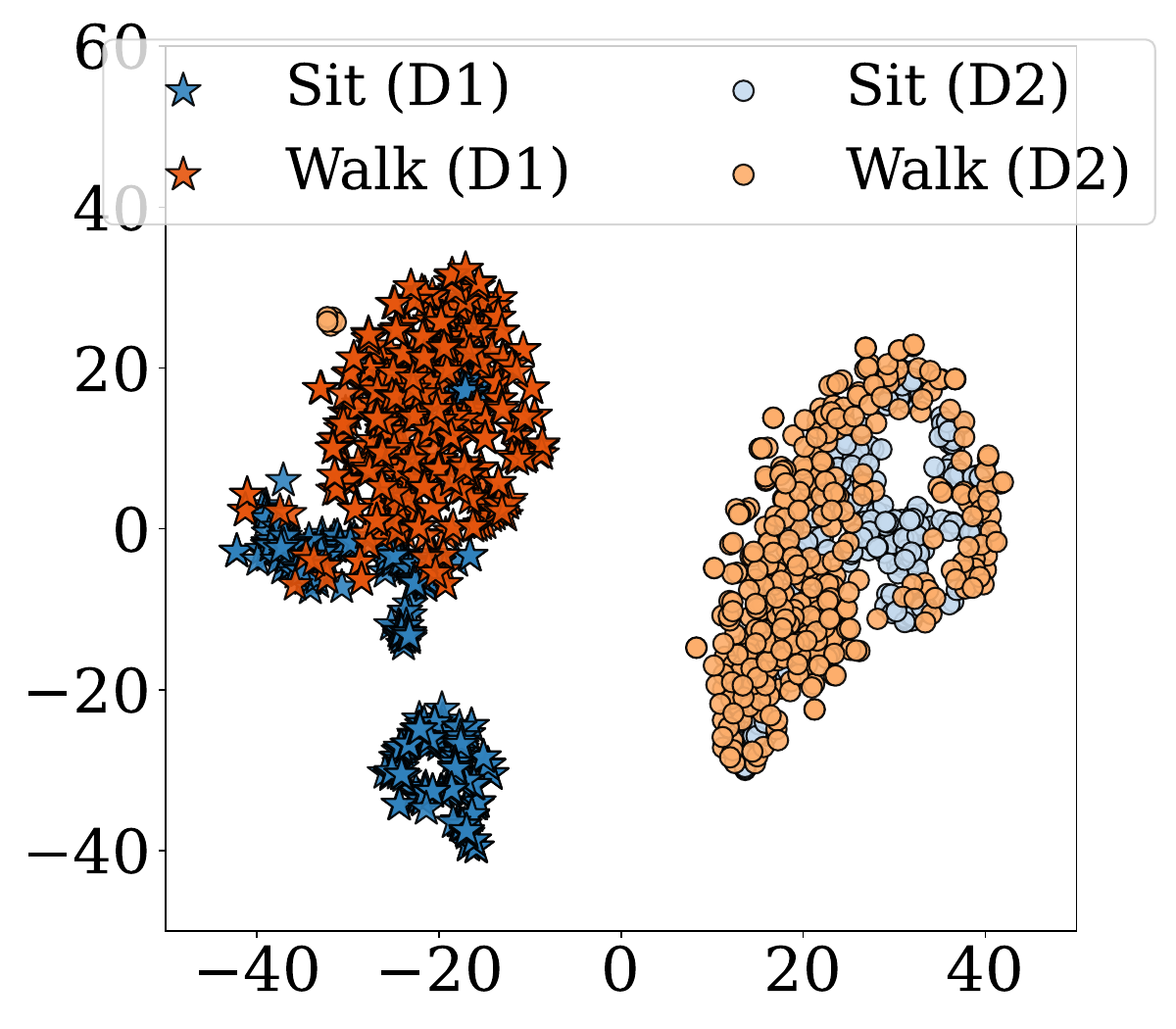}
        \vspace{-15pt}
        \caption{Raw data}
        \label{fig:raw data}
    \end{subfigure}
    \hspace{5pt}
    \begin{subfigure}{0.3\linewidth} \centering
        \includegraphics[width=\linewidth, keepaspectratio=true]{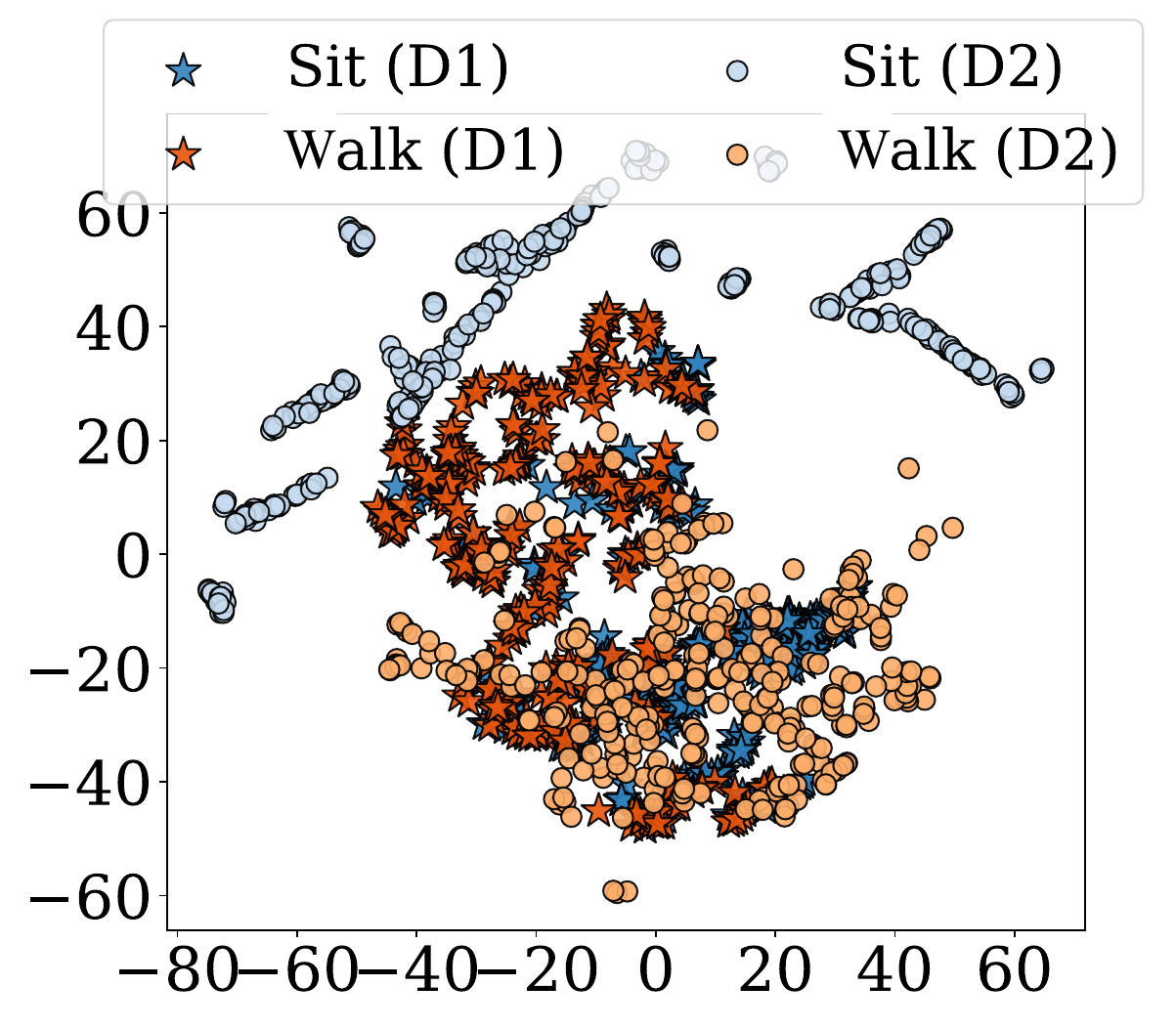}
        \vspace{-15pt}
        \caption{Self-supervised learning}
        \label{fig:self}
    \end{subfigure}
    \hspace{5pt}
    \begin{subfigure}{0.3\linewidth} \centering
        \includegraphics[width=\linewidth, keepaspectratio=true]{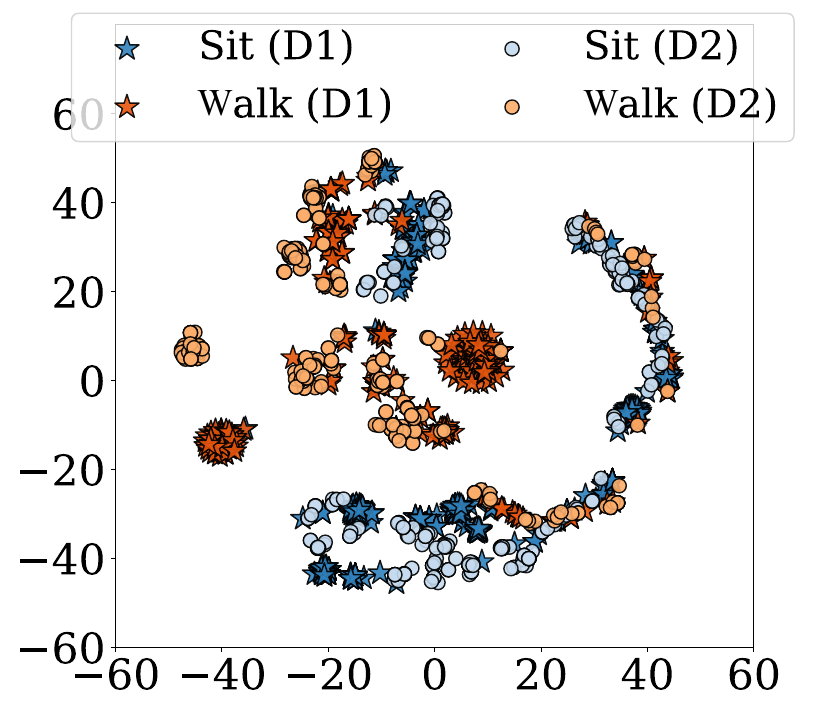}
        \vspace{-15pt}
        \caption{Semantic interpretations}
        \label{fig:Semantic}
    \end{subfigure}
    \vspace{-10pt}
        \captionsetup{font={normalsize}}
    \caption{Data distribution under three settings across two datasets}
    \label{fig:two distribution}
    \vspace{-10pt}
\end{figure*}

Existing research~\cite{xu2024penetrative,ji2024hargpt} has demonstrated that LLMs possess the ability to perceive and interpret the physical world. 
For instance, LLMs can analyze various types of sensor data to infer a person’s location or activity. 
Building on this capability, we explore the use of LLMs to analyze sensor readings and describe activity labels, thereby generating semantic interpretations for both. 
Specifically, we design prompts that include data introduction, data analysis, relevant knowledge, and task instructions (detailed in Section~\ref{sec: llm_prompt}) to guide LLMs in generating semantic interpretations of sensor readings. 
Similarly, we guide LLMs to describe activity labels. Fig.~\ref{fig:example} provides an example where GPT-4 generates its understanding of jogging alongside an analysis of sensor data sequences.

\subsection{Semantic interpretations of sensor readings for cross-dataset HAR}

We explain why LLM-generated semantic interpretation offers a valuable method to address data heterogeneity in cross-dataset HAR. 
The key intuition is that, despite variations in how heterogeneous data manifests, there are often underlying common activity patterns.
By identifying these shared patterns, we can map the original heterogeneous sensor data into a common shared space (i.e., a language space with encoded semantic interpretations), effectively reducing the impact of data heterogeneity.

To understand how data heterogeneity is mitigated using generated semantic interpretations, we compare data distributions under three settings: raw data, data encoded by encoders, and generated semantic interpretations. 
Data encoding is a common method in existing research~\cite{auge2021survey}, aimed at learning better representations or embeddings of sensor readings. 
For data encoding, we employ a self-supervised learning model, BERT~\cite{devlin2018bert}, as a representative method, following prior work~\cite{xu2021limu} that masks and predicts parts of the sensor readings. 
For semantic interpretation, we use GPT-4 to generate interpretations from raw IMU data based on a designed prompt (detailed in Section~\ref{sec: llm_prompt}). 
These interpretations are then input into a pre-trained language model (BERT) to produce representations in the language space. 
We use two public datasets—UCI~\cite{reyes2016transition} (D1) and Motion~\cite{malekzadeh2019mobile} (D2) as a cross-dataset example including four common activities: walking, sitting, going upstairs, and going downstairs.

\textbf{Qualitative results}: We visualize the results from the three settings using t-distributed Stochastic Neighbor Embedding (t-SNE)~\cite{van2008visualizing} to project them onto a 2D surface.
As an example, we plot two activities, ``sit'' (blue) and ``walk'' (orange), in Fig.~\ref{fig:two distribution}. 
Initially, in the raw data, the same activities from different datasets are far apart. 
After applying self-supervised learning, the distribution gap between the datasets decreases, but the activities also become less distinguishable. 
However, when using semantic interpretations, the distribution gap reduces further, and the same activities (same color) cluster more closely, demonstrating improved grouping of the same activities.

\textbf{Quantitative results}: 
We quantitatively assess the impact of introducing semantic interpretations for sensor readings across all activities.
To measure the distribution differences between the two datasets under the three settings, we use relative Kullback-Leibler (KL) divergence, computed as the ratio of KL divergence between the distributions of the same activity to KL divergence between the distributions of different activities~\cite{kullback1951information}, a widely used metric for comparing distributions. 
Compared to KL divergence, relative KL divergence normalizes differences in scale, mitigates the inherent closeness of word embeddings, and ensures a fair comparison across activities.
Table~\ref{tb: KL divergence} presents the relative KL divergence for all activities across the two datasets under each setting.
The results show that introducing semantic interpretations significantly reduces the distribution gap for all activities.
On average, the KL divergence is reduced by 40.45\%.

\begin{table}[t]
        \captionsetup{font={normalsize}}
\caption{Relative KL divergence between the distributions of the same activity across two datasets}
\label{tb: KL divergence}
\vspace{-10pt}
\resizebox{0.55\columnwidth}{!}{
\begin{tabular}{cccc}
\toprule
       \makecell{Activity}             & Raw data & \makecell{Self-supervised \\ learning} &  \makecell{Semantic \\ interpretations} \\ \midrule
Walking            &  0.4965    & 0.2331 & 0.1633     \\ 
Sitting             &  0.1669    & 0.1983 &  0.1541     \\ 
Going upstairs   &  0.2444     & 0.1875 &  0.0420     \\ 
Going downstairs &  0.1753     & 0.1518 &  0.0995     \\ \midrule
Average            &  0.2708     & 0.1926 &  0.1147     \\ 
\bottomrule
\end{tabular}
}
\vspace{-10pt}
\end{table}

\subsection{New activity recognition}

Recognizing new activities has been a longstanding challenge in HAR, particularly for machine learning-based methods, which are unable to identify activities that were not present during the training phase. The most relevant method to address this challenge involves using predefined attributes of activities to establish connections between new and existing activities~\cite{cheng2013nuactiv}. 
However, the performance of this method relies heavily on the quality of manually defined activity attributes, making it costly and difficult to generalize to new labels.

\begin{figure}[h] \centering
        \includegraphics[width=0.52\linewidth, keepaspectratio=true]{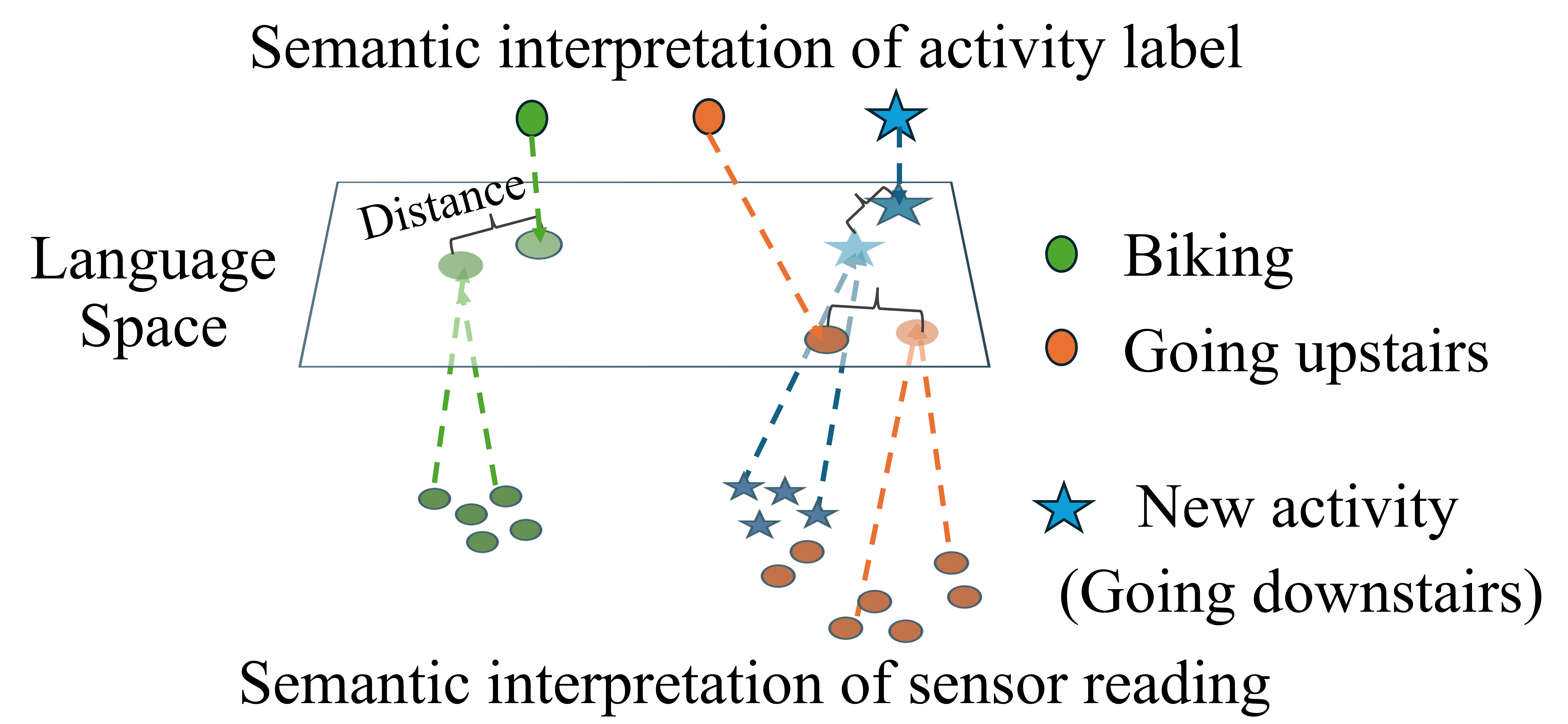}
        \vspace{-10pt}
        \caption{Semantic interpretations for new
activities}
        \label{fig:Sem_new}
        \vspace{-15pt}
\end{figure}

We argue that combining the semantic interpretations of sensor readings and labels offers a new perspective for addressing this challenge. 
By converting all activity labels into a language space using their semantic interpretations, we can represent labels even without corresponding sensor readings (which is common for new labels). 
Similarly, with sensor readings mapped into the same language space, we can match each reading to a label, regardless of whether the label was previously encountered. 
Fig.~\ref{fig:Sem_new} illustrates this concept, where we assume ``biking'' and ``going upstairs'' are known activities from the training phase, while ``going downstairs'' is a new activity.
In our method, we first convert all activity labels into the same language space using their LLM-generated semantic interpretations. 
We then obtain semantic interpretations for the sensor readings. 
Even though ``going downstairs'' is unseen during training, we can still measure its distance to all activity labels in the language space, enabling the recognition of new activities.
In our setting, we assume that we have the knowledge of possible activity labels in the inference stage.
It is important to note that our approach does not consider open-set recognition or arbitrary label discovery. Instead, we follow a closed-set assumption, where the candidate activity labels are known in advance during inference. To support this, we pre-stored all semantic interpretations of activity labels and use them to guide activity recognition during inference. Note that this does not violate our ``new activity'' objective where we do not have access to the training data of new activities, instead, simply the ``name'' of the activities.

\section{System design}

\subsection{Problem formulation}
We consider \name\ to be implemented in a cloud-client setting, where each client possesses locally collected, unlabeled IMU datasets.
The cloud holds an initial source dataset, $\textbf{X}^{s}$, collected from a subset of clients, with labels $\textbf{Y}^{s}$ provided by clients or experts.
Given an unlabeled target dataset {$\textbf{X}^{t}$} from the clients, our goal is to achieve high activity recognition accuracy on {$\textbf{X}^{t}$}, even when the label set $\textbf{Y}^{s}$ differs from $\textbf{Y}^{t}$.

\begin{figure*}[t] \centering
        \includegraphics[width=0.95\linewidth, keepaspectratio=true]{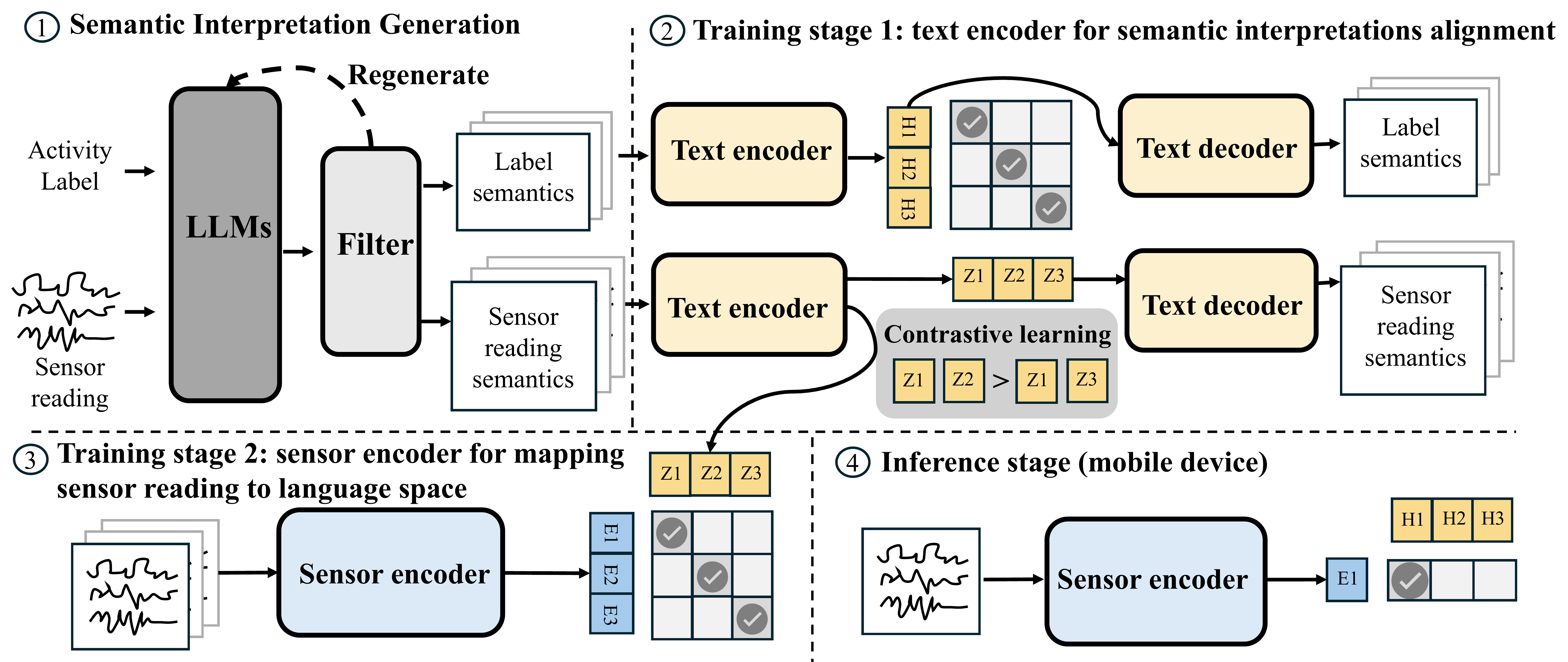}
        \vspace{-5pt}
        \captionsetup{font={normalsize}}
        \caption{Overview of \name{}. (1) Utilize LLMs for generating semantic interpretations of sensor reading and activity labels. (2) Train a text encoder to encode two types of semantic interpretations and achieve their alignment. $H{i}$ and $Z_{i}$ denote embeddings of semantic interpretations of activity labels and sensor reading. (3) Train a sensor encoder to align sensor reading and semantic interpretation. $E{i}$ denote embeddings of sensor reading. (4) For inference, only use the sensor encoder to generate embeddings for the sensor readings $E{i}$ and then compute the similarity with the pre-stored embeddings of the activity labels $H{i}$ to obtain the human activity recognition results.}
        \label{fig:overview}
        \vspace{-10pt}
\end{figure*}
\subsection{Design overview}
We design \name, a novel system that
uses LLM-generated semantic interpretations of sensor readings and activity labels for cross-dataset human activity recognition.  
As shown in Fig.~\ref{fig:overview}, \name\ consists of four processes:

\noindent\textbf{(1) LLMs for semantic interpretations.} We utilize LLMs to generate semantic interpretations of sensor readings $\textbf{X}^{s}$ and activity labels $\textbf{Y}^{s}$. We introduce an iterative re-generation process to filter out low-quality interpretations due to issues such as hallucinations.

\noindent\textbf{(2) Text encoder for semantic interpretations alignment.} We train a text encoder (i.e., initialized by a pre-trained language model) to encode and align the semantic interpretations generated in step (1). This alignment enables us to leverage language for human activity recognition by matching the semantic interpretation of sensor readings to the most similar semantic interpretation of activity labels within the language space.

\noindent\textbf{(3) Sensor encoder for mapping sensor reading to language space.} We train a sensor reading encoder, based on the text encoder from step (2), which is capable of mapping any sensor reading into the language space resulted from semantic interpretations.

\noindent\textbf{(4) Inference on the mobile device.} For inference, we deploy the sensor encoder from (3) on mobile devices to generate sensor embeddings,  which are then compared with pre-stored (new) activity label embeddings to determine the activity through similarity computation.

\subsection{LLMs for semantic interpretations}
\label{sec: llm_prompt}
We introduce how we guide LLMs to generate semantic interpretations of sensor readings and activity labels. We also discuss our method to handling low-quality interpretations to ensure the reliability of LLM responses.

\subsubsection{\textbf{Obtaining semantics interpretation of sensor reading}}
\label{sec: llm_prompt_reading}
\begin{figure}[t] \centering
        \includegraphics[width=0.54\linewidth, keepaspectratio=true]{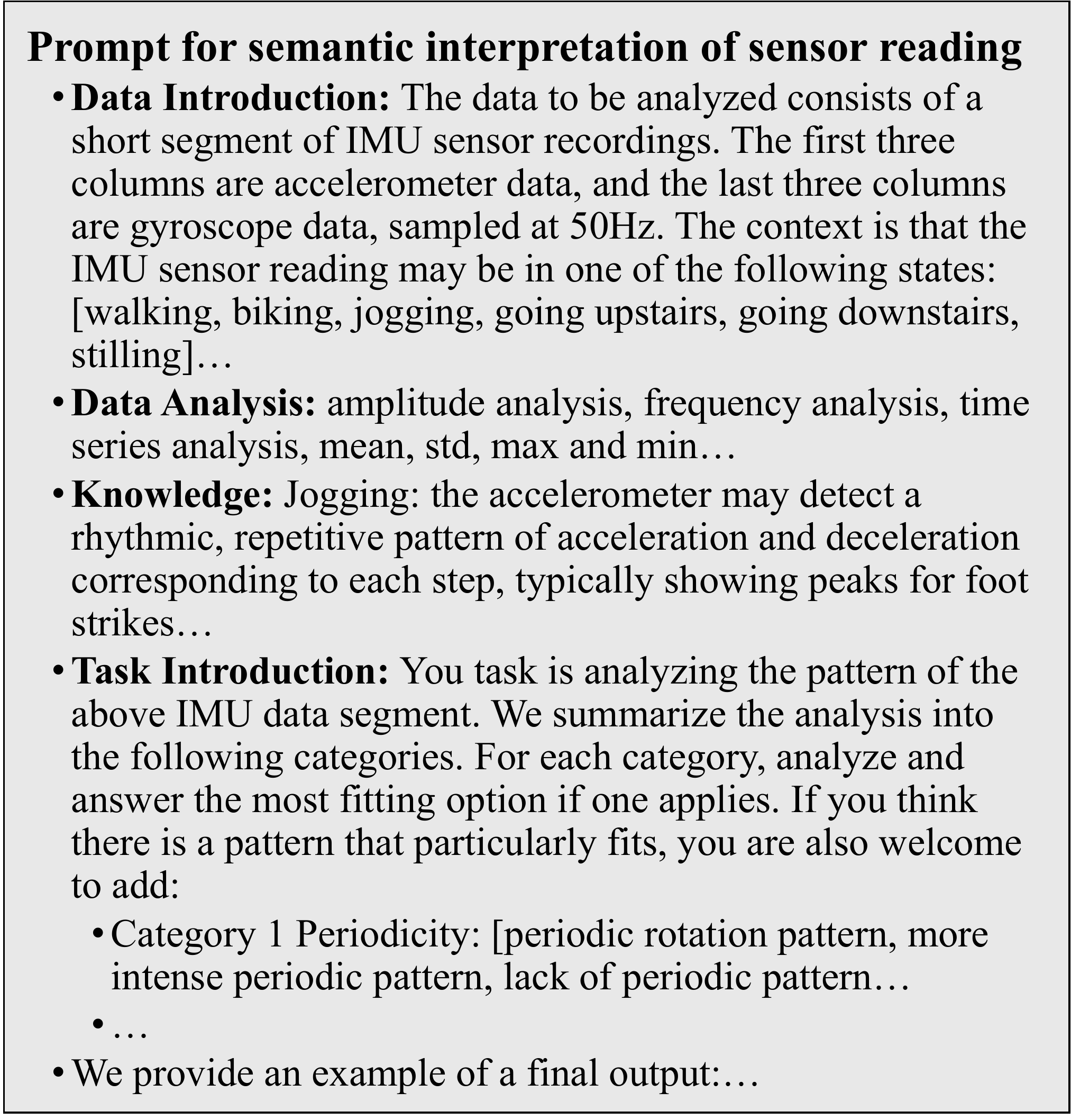}
        \captionsetup{font={normalsize}}
        \caption{An example prompt for obtaining semantic interpretation of sensor readings}
        \label{fig:prompt}
        \vspace{-10pt}
\end{figure}

We carefully design a prompt to help LLMs better understand the problem setting and deliver more accurate and meaningful answers. 
An example is shown in Fig.~\ref{fig:prompt}. The prompt consists of four key parts: data introduction, data analysis, knowledge, and task introduction.

\noindent\textbf{(i) Data introduction}: We provide details about the data, including its source, the meaning of each dimension, the sampling frequency, and the context in which the data is collected during different activities.

\noindent\textbf{(ii) Data analysis}:
Intermediate steps (i.e., Chain-of-Thought \cite{wei2022chain}) can enhance LLMs’ reasoning abilities. 
Our offline testing reveals that while LLMs are capable of performing various data analyses, skipping intermediate steps often leads to less confident responses. 
To address this, we design multiple auxiliary analysis steps (e.g., amplitude, frequency, time series analysis, and statistical measures such as mean, standard deviation, maximum, and minimum values). These steps help LLMs generate more accurate and precise answers.

\noindent\textbf{(iii) Knowledge}: 
We include relevant background knowledge to support the LLMs understanding the analysis procedure and focusing on the critical aspects among many possible features (e.g., impact peaks and vertical–horizontal dominance).
The background knowledge encompasses a few key aspects of sensor pattern characteristics and highlights their importance in distinguishing between different specific activities or broader activity categories.
The background knowledge is mainly derived from established findings in prior research on IMU-based human activity recognition~\cite{kavanagh2008accelerometry, nurwulan2021comparative, chen2021improving,wang2016comparative,barna2019study,li2009accurate,wang2011recognizing}. For example, we referred to studies on how vertical and horizontal components reflect different movement mechanisms, how inertial signals vary across activities, and how dominant gyroscope axes can indicate trunk rotation or arm-swing patterns. When needed, we also consulted open sources (e.g., Wikipedia) for basic terminology. All information was then consolidated through a combination of LLM-based summarization and manual refinement.

\noindent\textbf{(iv) Task introduction}: We outline the task clearly, specifying what the LLMs need to achieve and the format of the final output. This ensures that the LLMs have a clear understanding of the task requirements and can deliver standardized responses that meet our expectations.

\subsubsection{\textbf{Obtaining semantics interpretation of activity label}}
\begin{figure}[t] \centering
        \includegraphics[width=0.55\linewidth, keepaspectratio=true]{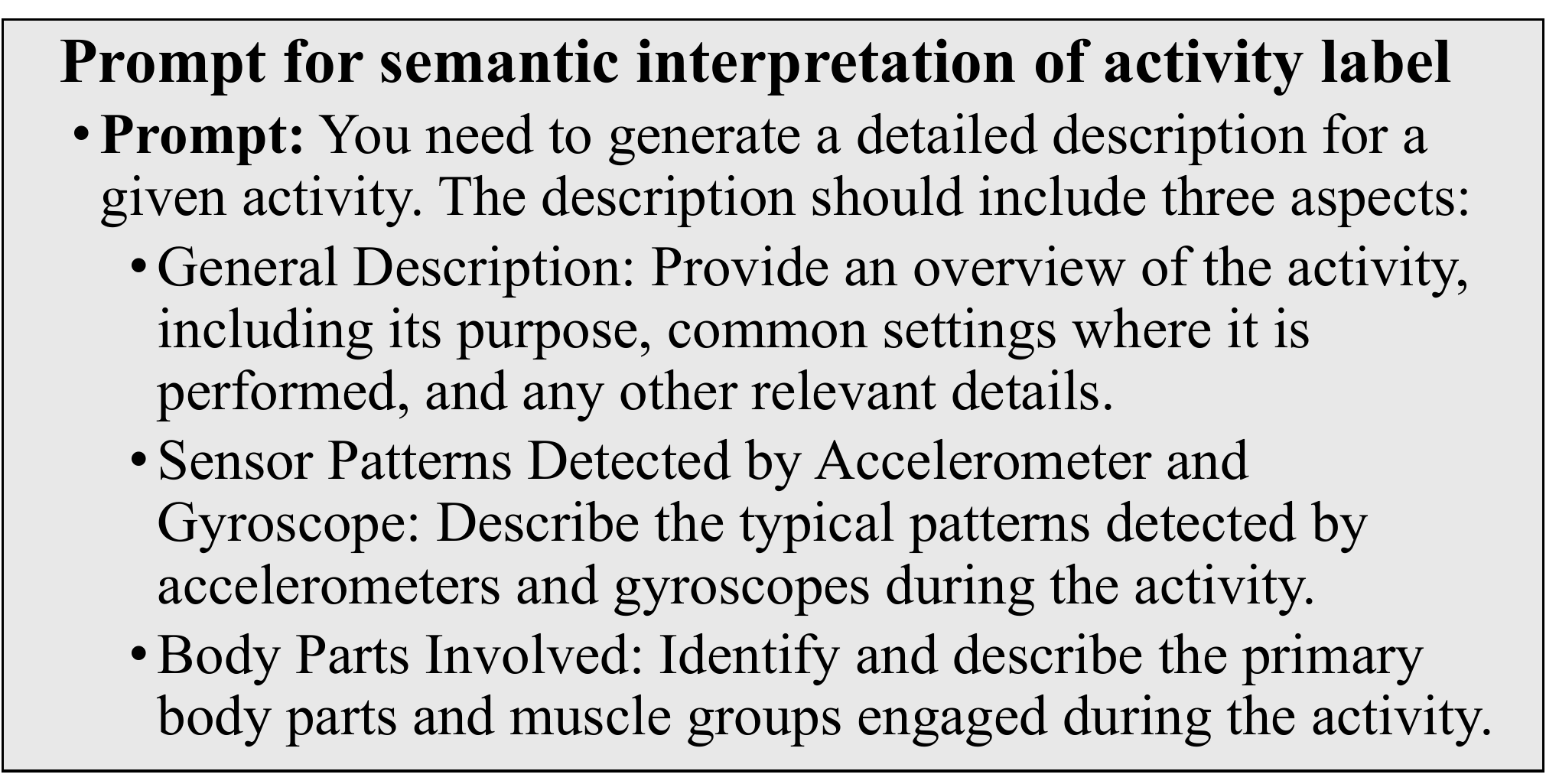}
        \captionsetup{font={normalsize}}
        \caption{An example prompt for obtaining semantic interpretation of activity labels}
        \label{fig:prompt_labels}
        \vspace{-10pt}
\end{figure}

Similarly, we design a prompt to guide LLMs in generating semantic interpretations of activity labels. In this prompt, we outline the task for the LLMs, instructing them to generate a description for a given activity. The description should cover three key aspects: a general overview of the activity, the potential states or patterns detected by the accelerometer and gyroscope during the activity, and the body parts likely involved in performing the activity. An example is shown in Fig.~\ref{fig:prompt_labels}.

\subsubsection{\textbf{An iterative re-generation method to ensuring the quality of LLM responses}}
\begin{figure}[t] \centering
        \includegraphics[width=0.5\linewidth, keepaspectratio=true]{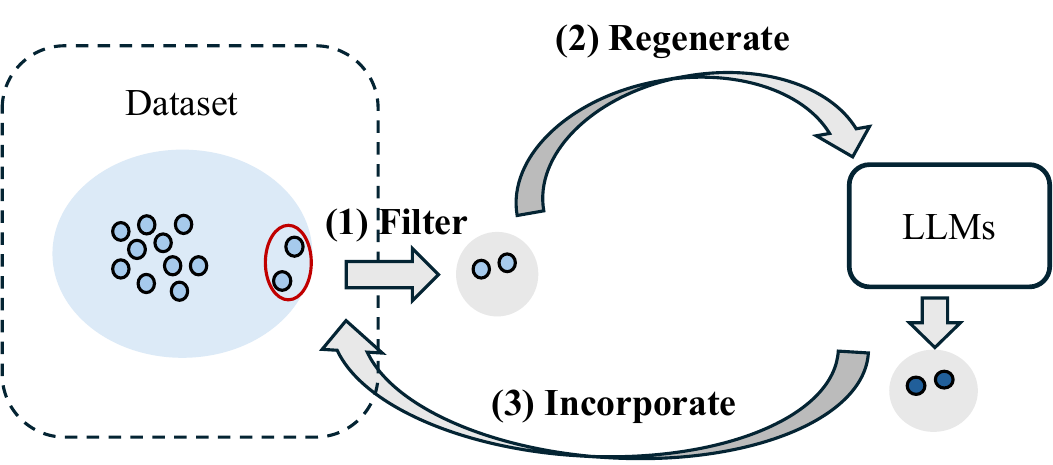}
        \vspace{-5pt}
        \captionsetup{font={normalsize}}
        \caption{The iterative re-generation method to ensuring the quality of LLM responses. (1) Filter inaccurate semantic interpretations. (2) Regenerate new semantic interpretations by LLMs. (3) Incorporate the new semantic interpretation based on its impact on the intra-cluster distance.}
        \label{fig:iteration}
        \vspace{-15pt}
\end{figure}

The quality of LLMs’ interpretations is critical for our method. 
However, due to hallucinations or other incidental factors, LLMs may sometimes generate inaccurate or illogical responses.  
Through manual review, we identify numerous instances of this issue when generating semantic interpretations.
To improve the accuracy of LLM responses, we develop an iterative re-generation method to ensure high-quality outputs.
The key intuition is that refining responses through clustering-based filtering will lead to a lower intra-cluster distance within the same activity.
As illustrated in Fig.~\ref{fig:iteration}, the framework first clusters the responses and identifies the $k$ most problematic ones based on their deviation from the cluster center.
These data points are re-input into the LLM to regenerate the semantic interpretations. We then assess whether to incorporate the new interpretations based on its impact on the intra-cluster distance.
The details are as follows. We show in the evaluation that this process improves the data quality.

\noindent\textbf{(1) Filter out inaccurate semantic interpretations:}
We first encode the generated semantic interpretations into dense vector representations using a pre-trained BERT encoder. We then apply a clustering-based method (i.e., K-means) to identify anomalous data points within the generated semantic interpretations of the same activity. After clustering, we select $k$ outliers based on their distance from the cluster center, with those farther than the average intra-cluster distance considered low-quality responses for refinement.

\noindent\textbf{(2) Regenerate new semantic interpretations by LLMs:} 
After identifying the $k$ outliers, we update the task introduction in the prompt (Section~\ref{sec: llm_prompt_reading}) to inform the LLMs of the inaccurate responses. 
The revised prompt includes the following: ``\textit{This is your previous response to the task. Please analyze it step by step to identify any logical errors, inconsistencies with real-world knowledge, or discrepancies with the input data. Provide a corrected response according to the required format}.''

\noindent\textbf{(3) Incorporate the new semantic interpretation based on its impact on the cluster structure:} If the newly generated interpretations reduce the intra-cluster distance, making the responses within the same cluster more compact, we accept the new semantic interpretation; otherwise, we retain the original one. We repeat this process until either the cluster structure remains stable (e.g., the intra-cluster distance change falls below a predefined threshold) for multiple continuous iterations or the specified number of iterations is reached.

Note that the iterative re-generation process occurs only during the offline training stage and does not impact the efficiency of online inference when deployed on mobile devices.
In our study, only 13.30\% of the semantic interpretations were identified as inaccurate.

\subsection{Text encoder for semantic interpretations alignment}
After generating the semantic interpretations of sensor readings and activity labels, we train a text encoder (initialized with a pre-trained language model) to encode and align these interpretations in a shared language space. To ensure the text encoder effectively captures the relationship between the semantic interpretations of sensor readings and activity labels, as well as their alignment, we design two subtasks: a contrastive learning task and a reconstruction task.

\subsubsection{\textbf{Text encoder}}
We adopt a pre-trained language model as the text encoder due to its robust ability to understand semantic interpretations through language. 
Specifically, we use the text encoder to encode the semantic interpretations of sensor readings $\mathbf{S}^{z}$ and activity labels $\mathbf{S}^{l}$, which are represented as $\mathbf{Z} = g_{\text{en}}(\mathbf{S}^{z})$ and $\mathbf{H} = g_{\text{en}}(\mathbf{S}^{l})$, where $g_{\text{en}}(\cdot)$ denotes the text encoder. 
In our work, we utilize the BERT architecture as the text encoder, initializing it with pre-trained BERT parameters~\cite{devlin2018bert}. 
BERT consists of multiple Transformer encoder layers, each incorporating multi-head self-attention mechanisms and feedforward neural networks, allowing it to generate word representations in a bidirectional context. 
We chose BERT because of its superior ability to understand text semantics and compare text similarity, as it captures word meanings in a bidirectional context, enabling nuanced comprehension and effective comparisons. Unlike generative models that process text in a left-to-right (or autoregressive) manner, BERT considers both the left and right context of each word simultaneously during training, allowing it to build deeper semantic representations based on the full sentence structure.
However, this text encoder structure is flexible and can be substituted with other language models if needed.

\subsubsection{\textbf{Alignment task}}
\label{sed:Alignment}

Inspired by multimodal alignment (e.g., vision and language)~\cite{radford2021learning}, we adopt contrastive learning to align the semantic representations of activity labels with those of sensor readings. 
This alignment enables the use of language to guide human activity recognition by matching the semantic interpretations of sensor readings to the most similar activity label interpretations in the language space.
First, we normalize both the embeddings of the semantic interpretations of sensor readings and activity labels to have unit length. 
Then, we compute the similarity between the normalized embeddings using the dot product, defined as:
\[
s(\mathbf{H},\mathbf{Z}) = \frac{\mathbf{H}}{||\mathbf{H}||} \cdot \frac{\mathbf{Z}}{||\mathbf{Z}||}
\]
The goal of training is to maximize the similarity between matching pairs of semantic interpretations and minimize the similarity between non-matching pairs. Given a batch of $N$ matching pairs, the loss function is defined as:
\[
\mathcal{L}_{\text{align}} = -\log \frac{\exp(s(H_i, Z_i)/\tau)}{\sum_{j=1}^{N} \exp(s(H_i, Z_j)/\tau)} -\log \frac{\exp(s(H_i, Z_i)/\tau)}{\sum_{j=1}^{N} \exp(s(H_j, Z_i)/\tau)},
\]
where $\tau$ is a temperature parameter that scales the similarities, and $H_i$ and $Z_i$ represent the embeddings of the semantic interpretations of sensor readings and activity labels for the $i$-th matching pair.

\subsubsection{\textbf{Contrastive learning tasks}}
We design two contrastive learning subtasks to enable the text encoder to fully understand the knowledge related to human activity recognition.
Firstly, we use pre-defined category-level relations (details in Section~\ref{sec:category har}) to guide the comparison between the semantic interpretations of sensor readings and activity labels.
For semantic interpretations of activity labels, two activity labels in the same category have a higher similarity than two activity labels in the different categories. 
The loss function for this task is defined as 
\[
\mathcal{L}_{C_{a_{1}}} = -\sum_{\mathcal{P}_{a_{1}}}(ln\sigma(s({H_{i}^{c_{m}} \cdot H_{j}^{c_{m}}})- s({H_{i}^{c_{m}} \cdot H_{k}^{c_{n}}}))),
\]
\noindent where $\mathcal{P}_{a_{1}}$ represents comparison pairs among the semantic interpretations of activity labels. $H_{i}^{c_{m}}$ and $H_{j}^{c_{m}}$ denote the embeddings of two activity label samples $i$ and $j$, that belong to category $c_{m}$.
Similarly, we apply the same comparison method to the semantic interpretations of sensor reading. 
The loss function for this task is defined as 
\[\mathcal{L}_{C_{a_{2}}} = -\sum_{\mathcal{P}_{a_{2}}}(ln\sigma(s({Z_{i}^{c_{m}} \cdot Z_{j}^{c_{m}}})- s({Z_{i}^{c_{m}} \cdot Z_{k}^{c_{n}}}))),\]
where $\mathcal{P}_{a_{2}}$ represents comparison pairs among the semantic interpretations of sensor readings.
$Z_{i}^{c_{m}}$ and $Z_{j}^{c_{m}}$ denote the embeddings of two semantic interpretations of sensor readings samples $i$ and $j$, that belong to category $c_{m}$.

Secondly, to enrich the semantic interpretations of activity labels and enable the model to adapt to diverse descriptions, we generate multiple descriptions for each activity label. 
These descriptions guide the comparison, ensuring that the similarity of descriptions for the same activity is higher than that for different activities.
The loss function for this task is defined as $\mathcal{L}_{C_{a_{3}}} = -\sum_{\mathcal{P}_{a_{3}}}(ln\sigma(s({H^{l_{m}}_{i} \cdot H^{l_{m}}_{j}})- s({H^{l_{m}}_{i} \cdot H^{l_{n}}_{j}})))$,
where $\mathcal{P}_{a_{3}}$ represents comparison pairs among the semantic interpretations of activity labels. $H^{l_{m}}_{i}$ and $H^{l_{m}}_{j}$ denote the embeddings of two different semantic interpretations of the activity label $l_{m}$.

\subsubsection{\textbf{Reconstruction task}}
Lastly, we design a reconstruction task to retain the characteristics of the language model, ensuring it can function as a language model to understand the new activity descriptions and new sensor pattern descriptions.
We design a text decoder to reconstruct semantic interpretations of sensor reading and activity labels. 
The objective of the text decoder can be formulated as $\mathbf{\hat{S}^{z}} = g_{\text{de}}(\mathbf{Z})$ and $\mathbf{\hat{S}^{l}} = g_{\text{de}}(\mathbf{H})$, where $\mathbf{Z}$ and $\mathbf{H}$ denote embeddings of semantic interpretations of sensor readings and activity labels, respectively, and $\mathbf{\hat{S}^{z}}$ and $\mathbf{\hat{S}^{l}}$ denote reconstructed semantic interpretations of sensor reading and activity labels, respectively.
The text decoder consists of multiple layers of the Transformer decoder, each containing multi-head attention mechanisms and feedforward neural networks.
The loss function of the reconstruction task is defined as
$\mathcal{L}_{\text{re}} = \frac{1}{N}\sum_{i}^{N} CE(\hat{S}^{z}_{i}, S^{z}_{i})+\frac{1}{N}\sum_{i}^{N} CE(\hat{S}^{l}_{i}, S^{l}_{i})$, where CE denotes Cross-Entropy (CE) loss and $N$ denotes the number of training samples.
In total, the loss function of training the text encoder is defined as 
\[    \mathcal{L}_{\text{text}} = \mathcal{L}_{\text{align}} + \alpha(\mathcal{L}_{C_{a_{1}}} +\mathcal{L}_{C_{a_{2}}} + \mathcal{L}_{C_{a_{3}}}) + \beta(\mathcal{L}_{\text{re}}),\]
where $\alpha$ and $\beta$ are weighting factors that control the contributions of the different tasks.

\subsection{Sensor encoder for mapping sensor reading to language space}
To bridge IMU sensor readings with their semantic interpretations and enable mobile deployment, we introduce a sensor encoder that maps IMU sensor readings to the language space, aligning them with the semantic interpretation encodings of the same sensor readings. 
We first describe the structure of the sensor encoder, followed by an explanation of the training process.

\subsubsection{\textbf{Sensor encoder structure}}

The objective of the sensor encoder is to capture key features and generate embeddings from IMU data, which can be formulated as:
$\mathit{\mathbf{E}} = f_{sen}(\mathbf{X})$
where $\mathbf{X} \in \mathbb{R}^{S_{\text{dim}} \times L}$ represents the IMU sensor readings, and $\mathbf{E}$ denotes the corresponding embeddings. The structure of the sensor encoder is flexible, allowing for either a CNN or Transformer-based architecture. In our work, we adopt a Transformer-based sensor encoder.
First, following~\cite{xu2021limu}, we normalize the IMU sensor readings $\mathbf{X}$ to obtain the normalized readings, $\mathbf{X^{nor}}$. Next, we project the normalized sensor readings through a linear layer and apply layer normalization to the projected data: $\mathbf{X^{'}} = \text{LayerNorm}(W_{p}\mathbf{X^{nor}})$
where $W_{p}$ is the weight matrix of the linear layer. Then, we add positional encoding to the normalized data:$\mathbf{X^{0}} = \mathbf{X^{'}} + \text{PositionEncoding}(\mathbf{X^{'}})$
The position-encoded data $\mathbf{X^{0}}$ is then fed through the Transformer encoder, which processes the data across multiple layers to capture complex relationships and dependencies. Each Transformer layer consists of two main sub-layers: Multi-Head Self-Attention and a Feed-Forward Network, both followed by layer normalization and residual connections.
For the $i$-th encoder layer, the operations are as follows:
First, the Multi-Head Self-Attention mechanism calculates self-attention scores, producing the output:
\begin{equation}
\small
\begin{aligned}[t]
    \mathbf{X^{A_{i}}} &= \text{Attention}(\mathbf{X^{i-1}}) \\
    \mathbf{X^{AN_{i}}} &= \text{LayerNorm}(\mathbf{X^{i-1}} + \mathbf{X^{A_{i}}})
\end{aligned}
\end{equation}
Next, the output from the Multi-Head Self-Attention sub-layer is passed through the Feed-Forward Network:
\begin{equation}
\small
\begin{aligned}
    \mathbf{X^{ANF_{i}}} &= \text{FeedForward}(\mathbf{X^{AN_{i}}})\\
    \mathbf{X^{i}} &= \text{LayerNorm}(\mathbf{X^{ANF_{i}}} + \mathbf{X^{AN_{i}}})
\end{aligned}
\end{equation}
This process is repeated across multiple encoder layers, each feeding its output to the next.

\subsubsection{\textbf{Sensor encoder training.}}

Similar to the alignment task in Section~\ref{sed:Alignment}, we adopt contrastive learning to align embeddings of sensor reading with the embeddings of semantic interpretation of sensor reading. 
First, the embeddings of both the sensor reading and its semantic interpretation are normalized to unit length. 
Their similarity is calculated using the dot product between the normalized embeddings, denoted as $    s(\mathbf{E},\mathbf{Z}) = \frac{\mathbf{E}}{||\mathbf{E}||} \cdot \frac{\mathbf{Z}}{||\mathbf{Z}||}.$
The goal of the training is to maximize the similarity between matching pairs while minimizing the similarity between non-matching pairs. 
Given a batch of N sensor reading and semantic interpretation of sensor reading pairs, the loss function is computed as:
\begin{equation}
\small
\mathcal{L}_{\text{sensor}} = -\log \frac{\exp(s(E_i, Z_i)/\tau)}{\sum_{j=1}^{N} \exp(s(E_i, Z_j)/\tau)} -\log \frac{\exp(s(E_i, Z_i)/\tau)}{\sum_{j=1}^{N} \exp(s(E_j, Z_i)/\tau)}
\end{equation}
where $\tau$ is a temperature parameter.

\subsection{Training and inference process}

\textbf{Training process:} (i) Training the text encoder: we use the semantic interpretations of sensor readings and activity labels from the source dataset to train the text encoder, embedding human activity recognition-related knowledge and semantics.
(ii) Training the sensor encoder for all datasets using raw IMU data and its semantic interpretations (i.e., unlabeled data): The sensor encoder is designed to map any sensor reading into the semantic interpretation space of sensor readings.

\noindent\textbf{Inference process:}
The model operates in an inference-only mode and does not require access to additional unlabeled data, as both the sensor encoder and activity label embeddings are pre-trained and deployed on the device. 
At inference time, IMU sensor readings are encoded and compared against the pre-stored activity label embeddings to compute similarity scores. The label with the highest similarity is selected as the recognized activity. This same process also applies when recognizing new activities.

\section{Evaluation}
\subsection{Evaluation setup}
\subsubsection{\textbf{Dataset}}
\label{sec:data}
We first evaluate the performance of \name\ using four public datasets with simple activities that have been extensively employed in existing research~\cite{xu2023practically}.
(1) HHAR~\cite{stisen2015smart} includes individuals performing six activities (sitting, standing, walking, going downstairs, going upstairs, and biking) using six different types of mobile phones. 	
(2) UCI~\cite{reyes2016transition} captures six activities (i.e., standing, sitting, lying, walking, going downstairs, and going upstairs).
(3) Motion~\cite{malekzadeh2019mobile} involves 24 participants performing six activities (sitting, standing, walking, going upstairs, going downstairs, and jogging). 
(4) Shoaib~\cite{shoaib2014fusion} includes 10 participants engaged in seven activities (sitting, standing, walking, going upstairs, going downstairs, jogging, and biking). 
In addition, we also select datasets (i.e., PAMAP2~\cite{pamap2_physical_activity_monitoring_231}) with more complex activities to show the performance of \name. PAMAP2 consists of data collected from 9 participants performing 12 different activities, including both simple activities such as sitting, standing, and walking, and more complex activities such as vacuum cleaning and ironing.

\subsubsection{\textbf{Data preprocessing}}
We preprocess all datasets to ensure consistent sampling rates and window sizes. 
The sampling rate is set to 50 Hz to simulate a resource-constrained scenario, following prior studies~\cite{xu2023practically}. 
For cross-dataset activity recognition, we focus on four common activities (walking, going upstairs, going downstairs, and still) that are shared across the four datasets. Following prior work~\cite{xu2023practically}, we group sitting and standing under the unified category still.
We select one dataset as the labeled source and one of the remaining three as the target, resulting in 12 source–target combinations. Datasets are categorized based on their roles as either source or target. For source datasets, each is split into a training set (80\%) and a validation set (20\%). For target datasets, the entire dataset is used as the test set to directly assess cross-dataset performance.
In addition, during the second stage of training, we incorporate unlabeled data from the remaining source datasets to jointly train the sensor encoder. This helps the model learn more generalizable representations, even though the main supervision still comes from a single labeled source.
For new activity recognition, we train the model on the four common activities from the source dataset. 
We then select activities from the target dataset that differ from the four common activities as new activities.
Finally, we evaluate the model's ability to recognize these new activities by determining which activity labels the new data correspond to.

\subsubsection{\textbf{Baselines}}
We compare our method \name\ with three categories of baselines: conventional HAR, cross-dataset HAR, and new activity baselines.

\noindent\textbf{(i) Conventional HAR baselines:} \textbf{DCNN~\cite{yang2015deep}} is widely used in the field of human activity recognition for their ability to automatically extract local features from time-series sensor data.
\textbf{Transformers~\cite{vaswani2017attention}} utilize self-attention mechanisms to capture long-range dependencies in sensor data. 
\textbf{LIMU-BERT~\cite{xu2021limu}} leverages masked language modeling and self-supervised learning to effectively utilize unlabeled IMU sensor data.

\noindent\textbf{(ii) Cross-dataset HAR baselines:} \textbf{SDMix~\cite{lu2022semantic}} 
It is a method for cross-domain activity recognition that uses the mixup method that addresses semantic inconsistencies caused by domain gaps.
\textbf{UDAHAR~\cite{chang2020systematic}} In our setting, we follow the feature alignment methods UDA paradigm, which aims to minimize the domain gap by explicitly aligning the feature distributions of the source and target domains.
\textbf{UniHAR~\cite{xu2023practically}} is a method for cross-dataset activity
recognition. 
It designs physical-informed augmentation methods to improve the cross-dataset HAR performance.
\textbf{UniMTS~\cite{zhang2024unimtsunifiedpretrainingmotion}} is a unified pre-training framework designed for motion time-series data.

\noindent\textbf{(iii) New activity baselines:}
\textbf{DCNN~\cite{yang2015deep}} 
To adapt it for recognizing new activities, we fine-tune it using a small portion of data from the target dataset.
\textbf{Nuactiv~\cite{cheng2013nuactiv}} is a method designed for new activities. It predefined activity label-related attributes: actions of a specific part of the body.
\textbf{UniMTS~\cite{zhang2024unimtsunifiedpretrainingmotion}} is a unified pre-training framework designed for motion time-series data, with zero-shot and few-shot generalization across diverse human activity recognition tasks. 
\textbf{HARGPT~\cite{ji2024hargpt}} leverages LLMs as zero-shot human activity recognizers by framing sensor-based activity recognition as a natural language processing task.

\noindent\textbf{(iv) Variants of our model:}
\textbf{(1) \name{} without Subtasks of Text Encoder (w/o STE).} We remove the subtasks from the text encoder and focus its training on alignment. 
\textbf{(2) \name\ without Training of Text Encoder (w/o TTE).} We remove the training part of the text encoder and directly use a pre-trained BERT model to replace it. 
\textbf{(3) \name\ with LLAMA2 (w LLAMA2).} We replace GPT-4 with LLAMA2 to obtain semantic interpretations.
\textbf{(4) \name\ with GPT-3 (w GPT-3).} We replace GPT-4 with GPT-3 to obtain semantic interpretations.
\textbf{(5) \name\ with Different Language Model (w DLM).} We use the transformer architecture to replace BERT in \name.
\textbf{(6) \name\ without Iterative re-Generation (w/o IG).} We remove the iterative re-generation method.
\textbf{(7) \name\ without Iterative re-Generation and designed Prompt (w/o IGP).} We remove the iterative re-generation method and the designed prompt. We directly use a simple prompt only containing data and task requirements.

\subsubsection{\textbf{Metric}}
We evaluate our model using two metrics: accuracy and F1 score. 
For cross-dataset HAR performance, we use accuracy to measure the correctness of the classification results for the four selected activities. 
For new activities, we use accuracy to evaluate the correctness of the classification results for the new activity only.

\subsubsection{\textbf{Implementation}}
We implement our method and baselines using PyTorch 1.9.0 in a Python 3.7 environment and train them on two NVIDIA RTX A5000 GPUs, each with 24 GB of memory. 
For LLMs, we choose GPT-4 as our LLMs to obtain semantic interpretations of sensor readings and activity labels.
For the text encoder, we use the BERT architecture as its structure and initialize it with the pre-trained BERT parameters~\cite{devlin2018bert}.
For the sensor encoder, we set the embedding dimension to $768$ to maintain consistency with pre-trained BERT. The multi-head attention parameter is set to $2$ and the encoder layer consists of $3$ layers.
We utilize the AdamW optimizer with a learning rate of 1e-5 and set the batch size to 32.
For iterative re-generation, we set the threshold for intra-cluster distance change to 0.0005.

\begin{table*}[t]
\small
        \captionsetup{font={normalsize}}
\caption{Overall performance. Bolded scores indicate the best, while underlined scores represent the second best.}
\setlength\tabcolsep{1.7pt}
\renewcommand{\arraystretch}{1}
\label{tb: overall performance}
\vspace{-10pt}
\begin{tabular}{c|c|cccccccccccc|c}
\toprule
\multirow{2}{*}{Method}      & Source & \multicolumn{3}{c|}{HHAR}                                       & \multicolumn{3}{c|}{UCI}                                         & \multicolumn{3}{c|}{Motion}                                   & \multicolumn{3}{c|}{Shoaib}                                  & \multirow{2}{*}{Average} \\ \cmidrule{2-14}
                             & Target & \multicolumn{1}{c}{UCI} & \multicolumn{1}{c}{Motion} & Shoaib & \multicolumn{1}{c}{HHAR} & \multicolumn{1}{c}{Motion} & Shoaib & \multicolumn{1}{c}{HHAR} & \multicolumn{1}{c}{UCI} & Shoaib & \multicolumn{1}{c}{HHAR} & \multicolumn{1}{c}{UCI} & Motion  & \\ \midrule
\multirow{2}{*}{DCNN}         & Accuracy    & \multicolumn{1}{c}{0.181}    & \multicolumn{1}{c}{0.220}       & 0.208     & \multicolumn{1}{c}{0.230}     & \multicolumn{1}{c}{0.122}       &  0.299      & \multicolumn{1}{c}{0.514}     & \multicolumn{1}{c}{0.605}    &  0.564      & \multicolumn{1}{c}{0.470}     & \multicolumn{1}{c}{0.651}    &  0.571       & 0.392 \\ 
                             & F1 scores    & \multicolumn{1}{c}{0.153}    & \multicolumn{1}{c}{0.136}       &  0.098     & \multicolumn{1}{c}{0.154}     & \multicolumn{1}{c}{0.069}       &  0.245      & \multicolumn{1}{c}{0.321}     & \multicolumn{1}{c}{0.424}    &   0.443     & \multicolumn{1}{c}{0.419}     & \multicolumn{1}{c}{0.338}    &  0.555      & 0.266 \\ 
\multirow{2}{*}{Transformer} &  Accuracy      & \multicolumn{1}{c}{0.465}    & \multicolumn{1}{c}{0.505}       & 0.387       & \multicolumn{1}{c}{0.512}     & \multicolumn{1}{c}{0.451}       & 0.400       & \multicolumn{1}{c}{0.428}     & \multicolumn{1}{c}{0.449}    &   0.423     & \multicolumn{1}{c}{0.506}     & \multicolumn{1}{c}{0.450}    &   0.401     & 0.449 \\ 
                             & F1 scores       & \multicolumn{1}{c}{0.318}    & \multicolumn{1}{c}{0.385}       & 0.224       & \multicolumn{1}{c}{0.169}     & \multicolumn{1}{c}{0.155}       &   0.143     & \multicolumn{1}{c}{0.256}     & \multicolumn{1}{c}{0.279}    & 0.314       & \multicolumn{1}{c}{0.168}     & \multicolumn{1}{c}{0.155}    &  0.143       & 0.231 \\ 
\multirow{2}{*}{LIMU-BERT} &  Accuracy      & \multicolumn{1}{c}{0.615}    & \multicolumn{1}{c}{0.257}       & 0.355       & \multicolumn{1}{c}{0.310}     & \multicolumn{1}{c}{0.315}       & 0.169      & \multicolumn{1}{c}{0.352}     & \multicolumn{1}{c}{0.292}    &   0.344     & \multicolumn{1}{c}{0.509}     & \multicolumn{1}{c}{0.392}    &   0.321     & 0.353 \\ 
                             & F1 scores       & \multicolumn{1}{c}{0.575}    & \multicolumn{1}{c}{0.116}       & 0.283       & \multicolumn{1}{c}{0.151}     & \multicolumn{1}{c}{0.191}       &   0.153     & \multicolumn{1}{c}{0.256}     & \multicolumn{1}{c}{0.173}    & 0.275       & \multicolumn{1}{c}{0.472}     & \multicolumn{1}{c}{0.281}    &  0.222       & 0.263 \\ \midrule
\multirow{2}{*}{UDAHAR}          & Accuracy       & \multicolumn{1}{c}{0.473}    & \multicolumn{1}{c}{0.481}       &  0.249      & \multicolumn{1}{c}{0.446}     & \multicolumn{1}{c}{0.417}       &  0.315      & \multicolumn{1}{c}{0.428}     & \multicolumn{1}{c}{0.559}    &   0.591     & \multicolumn{1}{c}{0.452}     & \multicolumn{1}{c}{0.551}    &   0.561     & 0.460 \\ 
                             &  F1 scores      & \multicolumn{1}{c}{0.317}    & \multicolumn{1}{c}{0.407}       &  0.191      & \multicolumn{1}{c}{0.279}     & \multicolumn{1}{c}{0.336}       &  0.292      & \multicolumn{1}{c}{0.355}     & \multicolumn{1}{c}{0.471}    &  0.473      & \multicolumn{1}{c}{0.265}     & \multicolumn{1}{c}{0.398}    &      0.281   & 0.339 \\ 
\multirow{2}{*}{SDMix}       & Accuracy       & \multicolumn{1}{c}{0.656}    & \multicolumn{1}{c}{0.694}       & 0.547      & \multicolumn{1}{c}{0.619}     & \multicolumn{1}{c}{0.567}       &   0.626     & \multicolumn{1}{c}{0.446}     & \multicolumn{1}{c}{0.531}    &   0.580     & \multicolumn{1}{c}{0.461}     & \multicolumn{1}{c}{0.549}    &  0.611       & 0.574 \\ 
                             & F1 scores       & \multicolumn{1}{c}{0.489}    & \multicolumn{1}{c}{\underline{0.657}}       & 0.369       & \multicolumn{1}{c}{0.653}     & \multicolumn{1}{c}{0.348}       &  0.542      & \multicolumn{1}{c}{0.195}     & \multicolumn{1}{c}{0.323}    &    0.448    & \multicolumn{1}{c}{0.205}     & \multicolumn{1}{c}{0.388}    &  0.400          & 0.423 \\ 
\multirow{2}{*}{UniMTS}       & Accuracy       & \multicolumn{1}{c}{0.623}    & \multicolumn{1}{c}{0.687}       & 0.542      & \multicolumn{1}{c}{0.604
}     & \multicolumn{1}{c}{0.542}       &   0.602     & \multicolumn{1}{c}{0.523}     & \multicolumn{1}{c}{0.598}    &   0.611     & \multicolumn{1}{c}{0.637}     & \multicolumn{1}{c}{0.505}    &  0.696      & 0.600 \\ 
                             & F1 scores       & \multicolumn{1}{c}{0.523}    & \multicolumn{1}{c}{0.501}       & 0.387       & \multicolumn{1}{c}{0.610}     & \multicolumn{1}{c}{0.351}       &  0.528      & \multicolumn{1}{c}{0.357}     & \multicolumn{1}{c}{0.436}    &    0.509    & \multicolumn{1}{c}{0.327}     & \multicolumn{1}{c}{0.402}    &  0.511          & 0.454 \\ 
\multirow{2}{*}{UniHAR}      &  Accuracy    & \multicolumn{1}{c}{\underline{0.739}}    & \multicolumn{1}{c}{\underline{0.752}}       & \underline{0.613}       & \multicolumn{1}{c}{\underline{0.805}}     & \multicolumn{1}{c}{\underline{0.786}}       &  \underline{0.741}      & \multicolumn{1}{c}{\underline{0.667}}     &  \multicolumn{1}{c}{\underline{0.735}}    &  \textbf{0.749}     & \multicolumn{1}{c}{\underline{0.525}}     & \multicolumn{1}{c}{\underline{0.767}}    & \textbf{0.817}   & \underline{0.721} \\ 
                             & F1 scores       & \multicolumn{1}{c}{\underline{0.604}}    & \multicolumn{1}{c}{0.651}       & \underline{0.592}       & \multicolumn{1}{c}{\underline{0.700}}     & \multicolumn{1}{c}{\underline{0.669}}       &  \underline{0.709}      & \multicolumn{1}{c}{\underline{0.581}}     & \multicolumn{1}{c}{\underline{0.681}}    & \underline{0.714}      & \multicolumn{1}{c}{\underline{0.480}}     & \multicolumn{1}{c}{\underline{0.681}}    & \textbf{0.777}    & \underline{0.646} \\ \midrule
\multirow{2}{*}{\name{}}        & Accuracy       & \multicolumn{1}{c}{\textbf{0.830}}    & \multicolumn{1}{c}{\textbf{0.812}}       &  \textbf{0.730}      & \multicolumn{1}{c}{\textbf{0.855}}     & \multicolumn{1}{c}{\textbf{0.811}}       &  \textbf{0.782}      & \multicolumn{1}{c}{\textbf{0.697}}     & \multicolumn{1}{c}{\textbf{0.764}}    &  \underline{0.730}      & \multicolumn{1}{c}{\textbf{0.717}}     & \multicolumn{1}{c}{\textbf{0.789}}    &  \underline{0.768}    & \textbf{0.774} \\ 
                             &  F1 scores     & \multicolumn{1}{c}{\textbf{0.804}}    & \multicolumn{1}{c}{\textbf{0.795}}       &  \textbf{0.693}      & \multicolumn{1}{c}{\textbf{0.808}}     & \multicolumn{1}{c}{\textbf{0.765}}       &   \textbf{0.718}     & \multicolumn{1}{c}{\textbf{0.617}}     & \multicolumn{1}{c}{\textbf{0.713}}    &    \textbf{0.724}    & \multicolumn{1}{c}{\textbf{0.687}}     & \multicolumn{1}{c}{\textbf{0.730}}    &    \underline{0.721}     & \textbf{0.731} \\ \bottomrule
\end{tabular}
\vspace{-10pt}
\end{table*}

\subsection{\textbf{Overall performance}}
\subsubsection{\textbf{Comparison to baselines using single labeled source dataset.}}
We compare our method on various baselines, and the results are presented in Table~\ref{tb: overall performance}. Overall, our method demonstrates superior average performance across 12 different source and target dataset combinations, leading in the majority of settings.
Compared to conventional HAR baselines, our method consistently outperforms them. These traditional baselines are not specifically designed for cross-dataset HAR, and their strong performance typically depends on large, carefully curated datasets.
Compared to cross-dataset HAR baselines, our method outperforms them (except in the setting from Shoaib to Motion).
Although SDMix and UDAHAR are also designed for cross-dataset HAR, they have specific requirements. 
For instance, UDAHAR may be more suitable for datasets with smaller distributional gaps. 
This consistent performance demonstrates the robustness and reliability of our method in diverse datasets.

Following the reviewer’s suggestion, we extend our cross-dataset experiments to include all relevant activities, treating sitting and standing as separate classes rather than merging them into a single still category. Specifically, the evaluated activities now include sitting, standing, walking, going upstairs, and going downstairs.
The new cross-dataset results for five activities are as shown in Appendix Table~\ref{tb: 5 activity performance}.
As shown in the results, the inclusion of both sitting and standing as separate activities introduces additional difficulty, as these two activities exhibit highly similar sensor patterns and are often hard to distinguish, especially in cross-dataset settings. This leads to a noticeable performance drop across all methods. While the performance of our method also drops due to the increased difficulty, it still achieves the best results among all compared approaches. This suggests that, despite the challenge of distinguishing highly similar activities such as sitting and standing, our method remains comparatively more effective.

\subsubsection{\textbf{Comparison to baselines using multiple labeled source datasets.}} 
In addition to baselines trained on a single labeled source dataset, some existing works aim to mitigate cross-dataset heterogeneity by leveraging multiple labeled source datasets. In this section, we compare our method with two types of approaches. The specific implementations are as follows: (i) GOAT~\cite{miao2024goat}: We implemented the GOAT framework following the original paper's setting, where the model is pre-trained on three datasets and evaluated on a separate target dataset. Since most public datasets do not provide sensor position annotations, we removed the device location descriptions to ensure consistent input formats across datasets. (ii) Combining Public HAR Datasets~\cite{presotto2023combining}: We implement the two base models used in this work (i.e., HART~\cite{ek2022lightweight} and DeepConvLSTM~\cite{ordonez2016deep}) and adopt the multi-source pre-training setup on three datasets. However, the original method assumes access to the target dataset for fine-tuning, while our method does not require any target-domain labels. To ensure a fair comparison, we evaluated their approach under two settings: (i) No fine-tuning: testing on the target dataset without fine-tuning, and (ii) Fine-tuning: using 5\% of labeled target data. Results are reported in Table~\ref{tab:pre-trained_results}.

From the results in the no fine-tuning scenario, we observe that merely using more data for pre-training does not guarantee improved performance. While aggregating multiple datasets can help reduce domain differences, effective performance typically requires an extensive and diverse set of source data. Otherwise, substantial distribution gaps between source and target domains still limit the generalization of pre-trained models.
In contrast, although fine-tuning significantly boosts performance, it relies heavily on labeled target-domain data, which is frequently unavailable in practical scenarios. Our proposed method achieves robust performance without requiring any labeled data from the target domain. 
This demonstrates superior adaptability, performing notably better than other methods without fine-tuning and comparably to methods that use fine-tuning.

\begin{table}[h]
\small
\caption{Comparison to baselines using multiple labeled source datasets. Bolded scores indicate the best.}
\label{tab:pre-trained_results}
\vspace{-10pt}
\begin{tabular}{l|l|cccc}
\toprule
Method & Metric & HHAR & Motion & UCI & Shoaib \\
\cmidrule(lr){1-6}
\multirow{2}{*}{Goat} & Accuracy & 0.431 & 0.549 & 0.704 & 0.535 \\
& F1 scores  & 0.156 & 0.262 & 0.573 & 0.402 \\
\multirow{2}{*}{\makecell[l]{Public-HART\\\textit{(No fine-tuning)}}} & Accuracy & 0.432 & 0.542 & 0.695 & 0.593 \\
& F1 scores  & 0.169 & 0.253 & 0.567 & 0.462 \\
\multirow{2}{*}{\makecell[l]{Public-DeepConv\\\textit{(No fine-tuning)}}}
 & Accuracy & 0.458 & 0.576 & 0.731 & 0.646 \\
& F1 scores  & 0.221 & 0.311 & 0.609 & 0.545 \\
\multirow{2}{*}{\makecell[l]{Public-HART\\\textit{(Fine-tuning)}}} & Accuracy & \textbf{0.798} & \textbf{0.803} & \textbf{0.830} & \textbf{0.862} \\
& F1 scores  & \textbf{0.731} & \textbf{0.764} & \textbf{0.779} & \textbf{0.830} \\
\multirow{2}{*}{\makecell[l]{Public-DeepConv\\\textit{(Fine-tuning)}}} & Accuracy & 0.792 & 0.762 & 0.822 & 0.767 \\
& F1 scores  & 0.673 & 0.697 & 0.757 & 0.703 \\
\cmidrule(lr){1-6}
\multirow{2}{*}{LanHAR} & Accuracy & 0.756 & 0.797 & 0.794 & 0.747 \\
& F1 scores  & 0.704 & 0.760 & 0.749 & 0.712 \\
\bottomrule
\end{tabular}
\vspace{-10pt}
\end{table}

\subsection{\textbf{New activity performance}}
\begin{figure}[h]\centering
\begin{minipage}[h]{0.37\linewidth}
    \includegraphics[width=\linewidth, keepaspectratio=true]{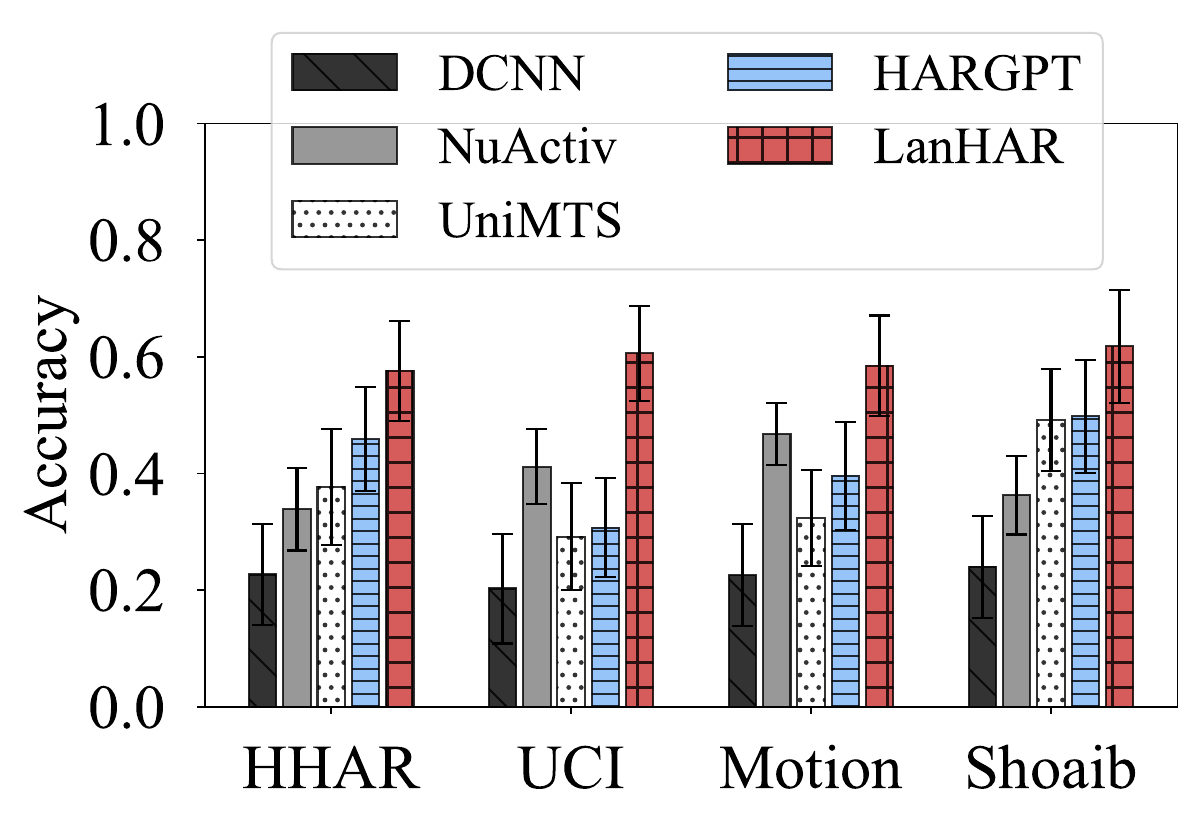}
    \vspace{-15pt}
        \captionsetup{font={normalsize}}
    \caption{New activity performance}
    \label{fig:New activity performance}
\end{minipage}
\hspace{10pt}
\begin{minipage}[h]{0.37\linewidth}
    \includegraphics[width=\linewidth, keepaspectratio=true]{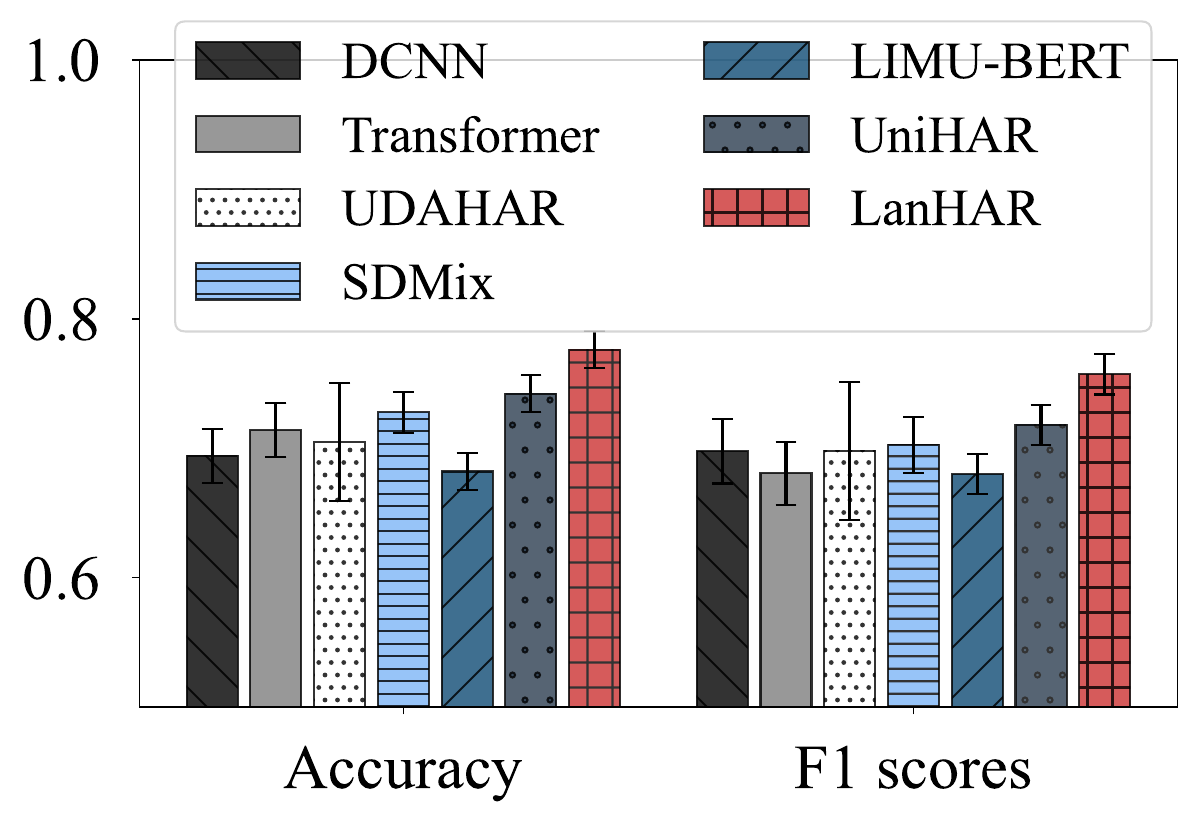}
    \vspace{-15pt}
        \captionsetup{font={normalsize}}
    \caption{Complex activity performance}
    \label{fig:Complex activity performance}
\end{minipage}
\vspace{-12pt}
\end{figure}

We compare our method with four new activity recognition baselines. Specifically, we train the model on four common activities from the source dataset and evaluate it on different activities from the target dataset that are not among the four common activities, treating them as new activities. The trained model is then used to recognize these new activities, i.e., determining which activity new data belong to. 
Fig.~\ref{fig:New activity performance} presents the performance of all methods in recognizing new activities across multiple target datasets.

For new activity recognition, our model outperforms all the baselines: 
Our model performance is superior to that of a fine-tuned DCNN model, which indicates its advantage over simple fine-tuning on small datasets. 
Additionally, our model performs better than Nuactiv, which relies on predefined attributes for activities. Defining these attributes is challenging and varies significantly across different activity contexts, directly impacting Nuactiv’s performance. 
This limitation hinders Nuactiv’s performance in cross-dataset settings.
Similarly, HARGPT, as a prompt-based HAR method, has limited ability to recognize new activities. UniMTS, although trained on large-scale simulated data, still does not perform well in zero-shot scenarios on real-world datasets.

\subsection{\textbf{Complex activity performance}}
In addition to simple activities such as sitting and walking, we also select datasets containing more complex activities such as vacuum cleaning, ironing, and rope jumping to validate the effectiveness of our model.
However, since different datasets contain different complex activities, to our knowledge, there are no two truly comparable datasets for cross-dataset HAR evaluation. 
Therefore, instead of using two different datasets, we select a single dataset and split it based on individuals to simulate cross-dataset performance.
Specifically, we randomly partition the PAMAP2 dataset into two subsets based on individuals’ IDs and select the partition with the highest KL divergence, indicating the greatest data variance across two subsets, as the final partitioning result.
The evaluation result is presented in Fig.~\ref{fig:Complex activity performance}. 
As illustrated, our model demonstrates superior performance compared to other models on datasets that include complex activities, achieving an improvement in accuracy of at least 4.58\% and at most 13.78\%, and an increase in F1 scores of at least 5.43\% and at most 11.32\%.

\subsection{\textbf{Non-cross-dataset performance}}
To evaluate the model's performance in a standard within-dataset setting, we conduct additional non-cross-dataset experiments. In this setup, the model is trained and tested on the same dataset (i.e., HHAR, Motion, UCI, and Shoaib). Specifically, each dataset is split into training and testing sets using an 8:2 ratio. The results are as shown in Table~\ref{tb: single performance}.

\begin{table}[h]
\small
        \captionsetup{font={normalsize}}
\caption{Non-cross-dataset performance. Bolded scores indicate the best.}
\setlength\tabcolsep{3.7pt}
\renewcommand{\arraystretch}{1}
\label{tb: single performance}
\vspace{-10pt}
\begin{tabular}{c|c|cccc}
\toprule
Method                       & Metric    & HHAR & UCI & Motion & Shoaib \\ \midrule
\multirow{2}{*}{DCNN}        & Accuracy  & \textbf{0.995}& 0.947& 0.980  & \textbf{0.982} \\ 
                             & F1 scores &\textbf{0.993}& 0.954& 0.967 & \textbf{0.982} \\
\multirow{2}{*}{Transformer} & Accuracy  & 0.932 & 0.924& 0.926 & 0.940\\ 
                             & F1 scores & 0.929&0.918 & 0.903 & 0.935 \\ 
\multirow{2}{*}{LIMU-BERT}    & Accuracy  & 0.961 & \textbf{0.976} & \textbf{0.983} & 0.940 \\ 
                             & F1 scores & 0.961 & \textbf{0.962} & 0.980 & 0.947 \\ \midrule
\multirow{2}{*}{UDAHAR}      & Accuracy  & 0.992 & 0.930 & 0.980 & 0.978 \\ 
                             & F1 scores & 0.991 & 0.938 & 0.967 &0.978\\ 
                             
\multirow{2}{*}{SDMix}       & Accuracy  & 0.949 & 0.957 & \textbf{0.983} & 0.924 \\ 
                             & F1 scores & 0.949 & 0.934 & \textbf{0.982} & 0.931\\ 
\multirow{2}{*}{UniMTS}      & Accuracy  & 0.936  & 0.935  &  0.953   & 0.961       \\ 
                             & F1 scores & 0.912  & 0.924     & 0.947  & 0.954       \\ 
\multirow{2}{*}{UniHAR}      & Accuracy  & 0.927 & 0.937 & 0.971 & 0.900\\ 
                             & F1 scores & 0.886 & 0.904 & 0.949 & 0.882\\ \midrule
\multirow{2}{*}{\name}      & Accuracy  & 0.956     & 0.951 &0.922 & 0.967 \\ 
                             & F1 scores & 0.949 &0.954 &  0.906 & 0.965  \\ \bottomrule
\end{tabular}
\vspace{-5pt}
\end{table}

As shown in the results, most methods, including ours, achieve high accuracy in this setting, likely because each single dataset is collected from a relatively homogeneous group of participants under similar conditions, resulting in low distributional variability.
Although our model is not the best-performing method on every dataset, it remains highly competitive. In terms of accuracy, the best-performing method on each dataset outperforms the worst-performing one by an average of 7.17\%, while the gap between the best-performing method and ours is only 3.72\% on average. Similarly, for F1-score, the average difference between the best and worst methods per dataset is 9.05\%, while the best-performing method exceeds ours by just 3.91\% on average.
The slightly lower performance of our model may be due to the loss of some dataset-specific information during the semantic interpretation process, which helps generalization across datasets but may limit performance within a single dataset.

\subsection{\textbf{Category-level HAR performance}}
\label{sec:category har}
\begin{figure}[h]\centering
\begin{minipage}[b]{0.37\linewidth}
    \includegraphics[width=\linewidth, keepaspectratio=true]{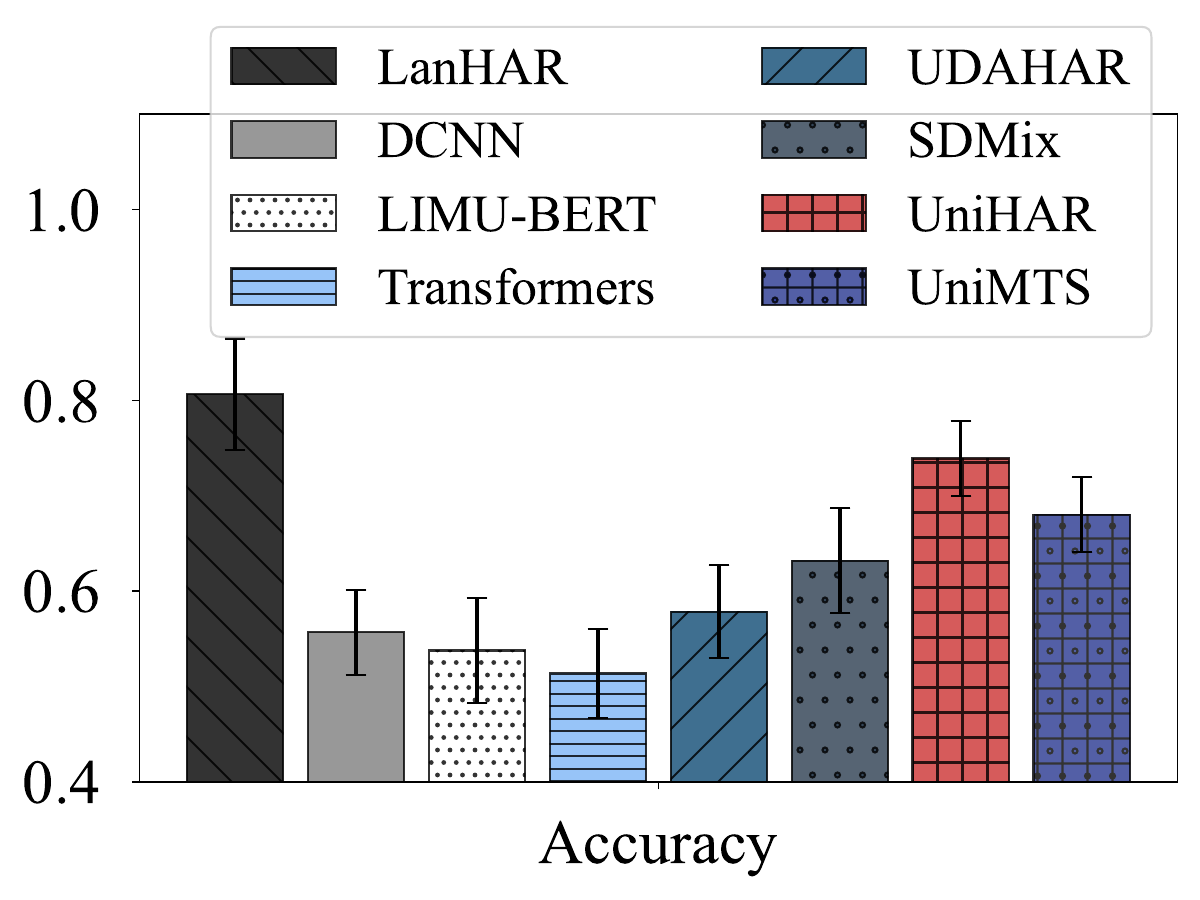}
    \vspace{-15pt}
        \captionsetup{font={normalsize}}
    \caption{Category-level HAR Performance}
    \label{fig:Category-level HAR Performance}
\end{minipage}
\hspace{10pt}
\begin{minipage}[b]{0.37\linewidth}
    \includegraphics[width=\linewidth, keepaspectratio=true]{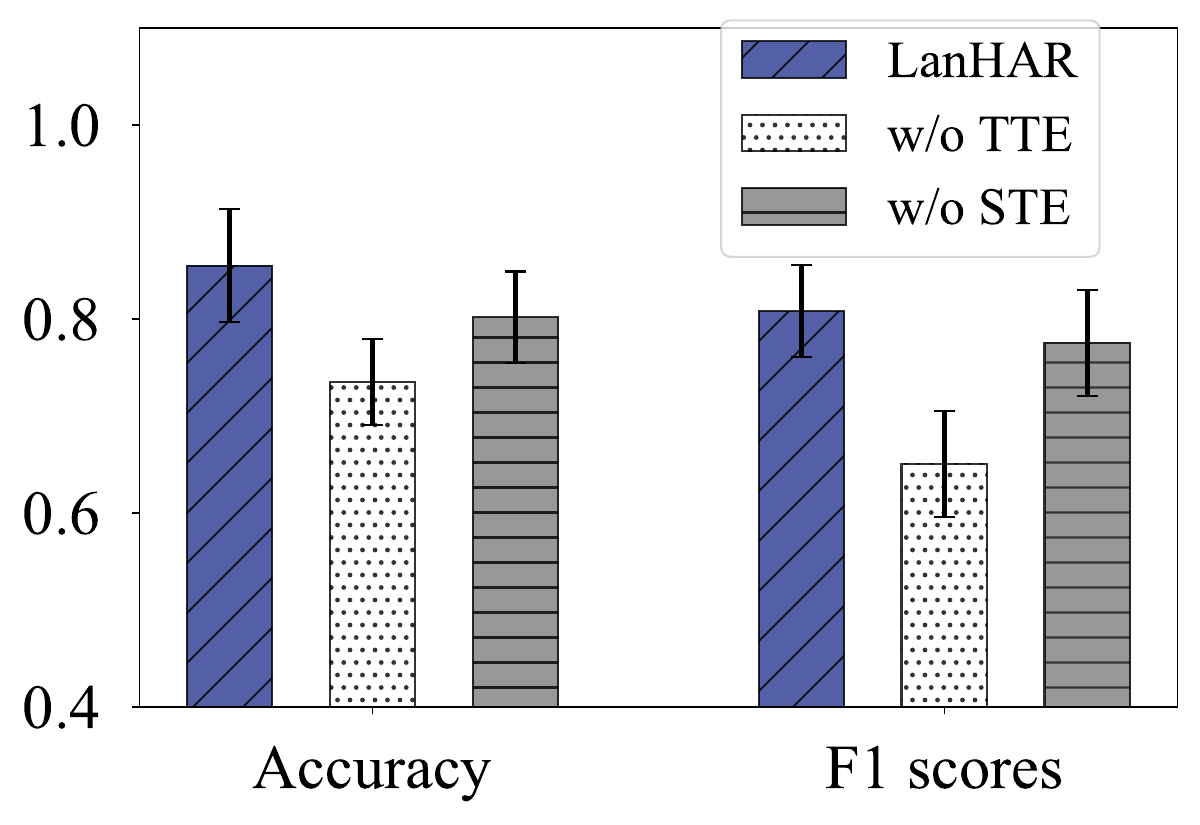}
    \vspace{-15pt}
        \captionsetup{font={normalsize}}
    \caption{The Effect of Text Encoder}
    \label{fig:effect of text encoder}
\end{minipage}
\vspace{-10pt}
\end{figure}

We further evaluate the category-level accuracy of HAR performance. An activity is considered correctly classified if it is assigned to a similar activity category and incorrectly classified if it falls into a dissimilar category. The motivation for this evaluation is that even if a model does not provide the correct prediction, we expect its misclassification to be a similar activity rather than a completely unrelated one.
Specifically, we group seven activities (walking, jogging, biking, sitting, standing, going upstairs, going downstairs, and lying) into three categories based on their similarity: Category 1 (walking, jogging, biking), Category 2 (sitting, standing, lying), and Category 3 (going upstairs, going downstairs).
We compare our method against all the baselines. 
The average results are shown in Fig.~\ref{fig:Category-level HAR Performance}. As shown, our method improves category-level HAR accuracy compared to the baselines. This improvement has the potential to reduce the impact of classification errors on downstream applications that require a broader activity categorization rather than fine-grained recognition.

\subsection{\textbf{Ablation study}}
\subsubsection{\textbf{The effect text encoder}}
To assess the impact of the subtasks in the text encoder, we compare our method with a variant (i.e., w/o STE), as shown in Fig.~\ref{fig:effect of text encoder}. 
In the w/o STE variant, we remove the subtasks (i.e., the two contrastive learning and reconstruction tasks) and focus only on aligning the semantic interpretations of sensor readings and activity labels. 
As shown in the figure, removing the subtasks leads to a decline in performance compared to our full model. This demonstrates that the subtasks we designed help the model better understand IMU sensor readings and HAR-related semantics.

To further explore the effect of the text encoder on semantic interpretation alignment, we replace the carefully designed text encoder with a pre-trained BERT model and directly use it for training in Stage 2. 
The results are shown in Fig.~\ref{fig:effect of text encoder}. 
From the figure, it is clear that the performance of w/o TTE is lower than both w/o STE and our full model. This suggests that pre-trained language models without fine-tuning cannot fully capture HAR-related semantics or effectively align different HAR-related semantic interpretations.

\subsubsection{\textbf{The impact of different pre-trained LMs}}
\begin{figure}[h]\centering
\begin{minipage}[h]{0.37\linewidth}
    \includegraphics[width=\linewidth, keepaspectratio=true]{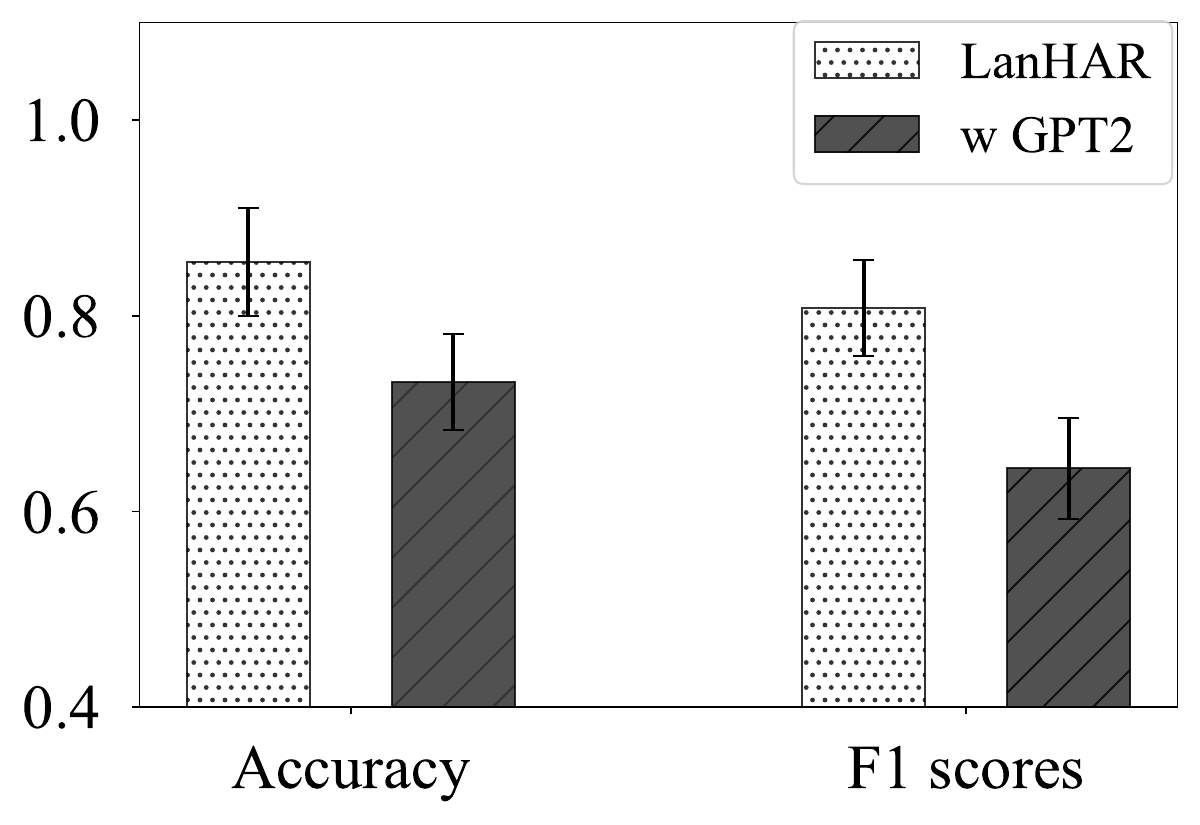}
    \vspace{-15pt}
        \captionsetup{font={normalsize}}
    \caption{The impact of different pre-trained Language Models as text encoders}
    \label{fig:effect of pre-trained LM}
\end{minipage}
\hspace{10pt}
\begin{minipage}[h]{0.37\linewidth}
    \includegraphics[width=\linewidth, keepaspectratio=true]{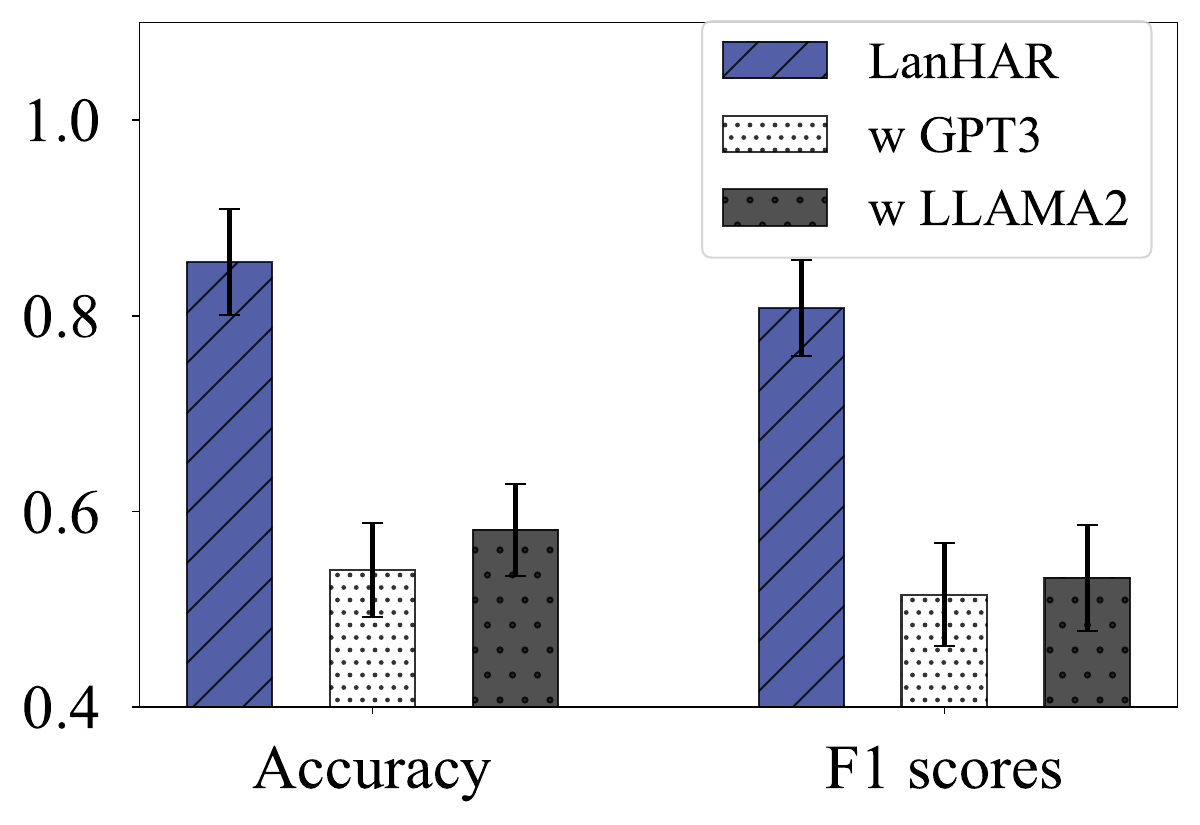}
    \vspace{-15pt}
        \captionsetup{font={normalsize}}
    \caption{The impact of different Large Language Models}
    \label{fig:The Effect of LLM}
\end{minipage}
\end{figure}

Our text encoder is flexible and allows the integration of different language models. In our work, we employ a BERT-based architecture, fine-tuned on our designed tasks. To investigate the impact of different language model architectures on HAR performance, we replace the BERT-based encoder with GPT-2.
As shown in Fig.~\ref{fig:effect of pre-trained LM}, this replacement results in decreased HAR performance.
BERT is generally more suitable for tasks that require understanding sentence meaning and comparing text similarities, as its architecture and pre-training are optimized for capturing semantics and contextual relationships in a bidirectional manner. 
In contrast, GPT-2, with its unidirectional processing, is less effective at understanding sentence-level meaning and comparing text similarity, though it excels in generating coherent and contextually relevant text. This explains the performance drop when GPT-2 is used for HAR tasks.

\subsubsection{\textbf{The impact of different LLMs}}
Given the varying capabilities of different LLMs, we explore how three LLMs (GPT-4, GPT-3, and LLAMA2) impact the interpretation of sensor readings and activity labels in HAR performance.
As shown in Fig.~\ref{fig:The Effect of LLM}, the performance of GPT-3 and LLAMA2 is noticeably lower than that of GPT-4, highlighting the importance of the LLM's capability in language-centered human activity recognition. These results suggest that stronger LLMs with advanced reasoning abilities are better suited for enhancing HAR performance.

\subsubsection{\textbf{The impact of different response qualities of LLMs }}

\begin{figure}[h]\centering
\begin{minipage}[h]{0.37\linewidth}
    \includegraphics[width=\linewidth, keepaspectratio=true]{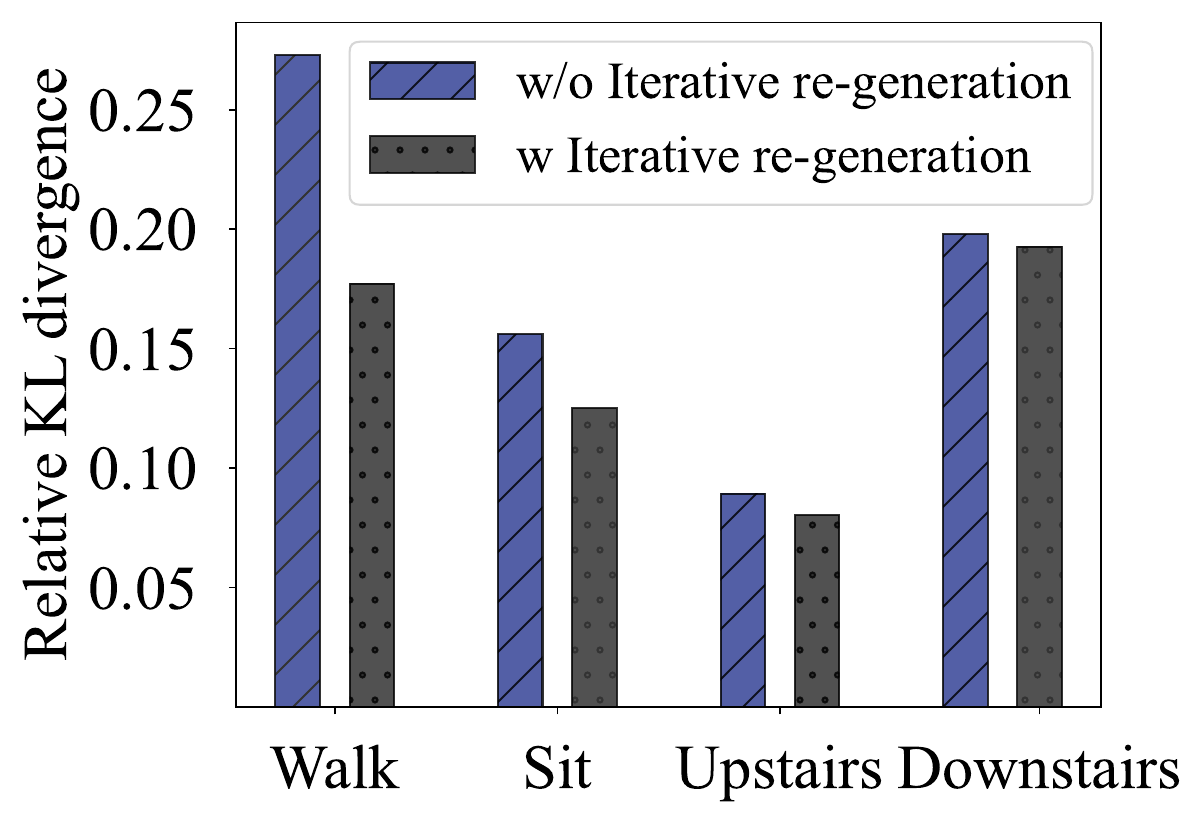}
    \vspace{-15pt}
          \captionsetup{font={normalsize}}
    \caption{Relative KL divergence with and without the iterative re-generation method}
    \label{fig:kl_change}
\end{minipage}
\hspace{10pt}
\begin{minipage}[h]{0.37\linewidth}
    \includegraphics[width=\linewidth, keepaspectratio=true]{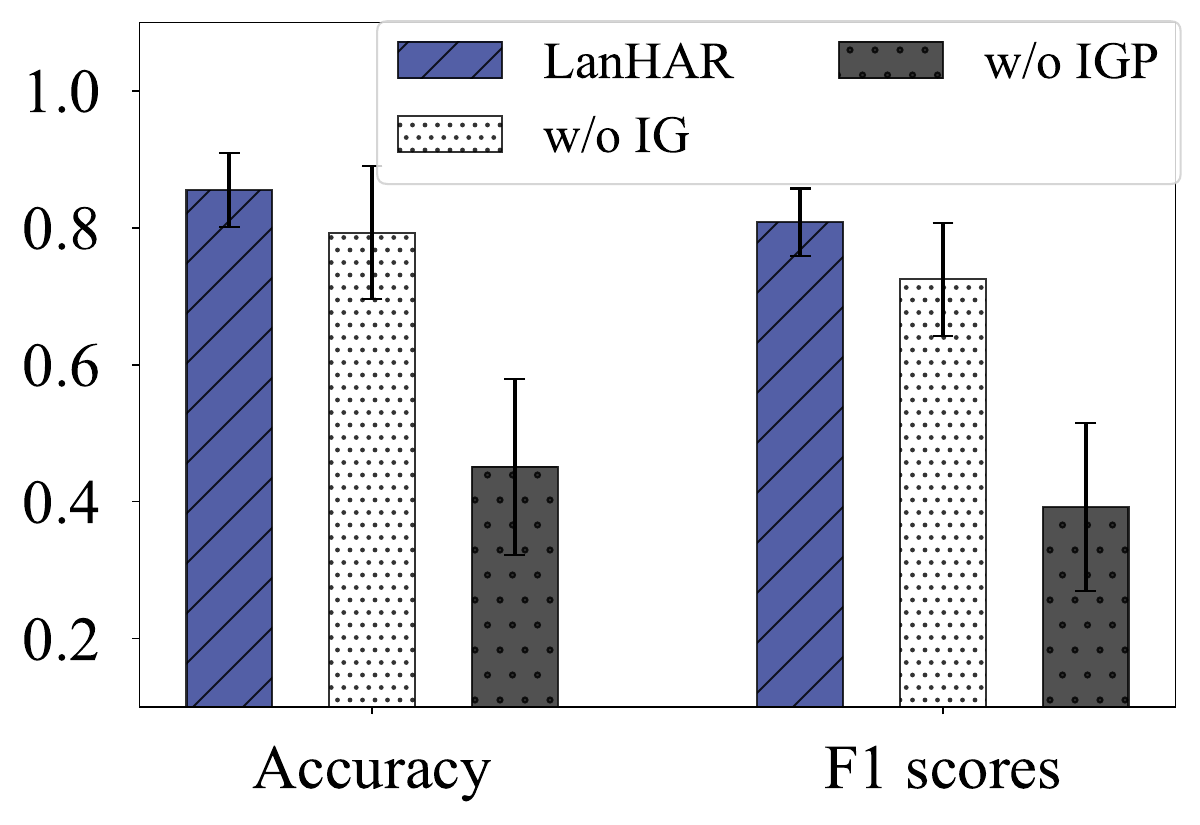}
    \vspace{-15pt}
   \captionsetup{font={normalsize}}
    \caption{The impact of different quality of LLMs’ response}
    \label{fig:quality of llm}
\end{minipage}
\end{figure}

To ensure the quality of the LLM’s interpretation of sensor readings, we (i) design specific prompts to guide the LLM’s responses and (ii) develop an iterative re-generation method to filter out low-quality semantic interpretations and generate improved ones.
To demonstrate the effectiveness of the automatic iterative re-generation method, we compare the relative KL divergence of the same activities across different datasets before and after applying the method. Specifically, for each activity, we measure the relative KL divergence between any two datasets both with and without applying the iterative re-generation method (i.e., ``w iterative re-generation'' and ``w/o iterative re-generation''), as shown in Fig.~\ref{fig:kl_change}. 
After applying the method, the average relative KL divergence for the same activity across different datasets decreases by 16.92\%, indicating that the automatic iterative re-generation method significantly improves the response qualities of LLMs and reduces the distribution gap in heterogeneous data.

To verify the impact of these methods on HAR performance, we compare our method with two variants: one using only simple prompts (w/o IGP) and another using only specifically designed prompts (w/o IG).
As shown in Fig.~\ref{fig:quality of llm}, the performance when using only simple prompts is the lowest, indicating that LLMs require guidance to generate high-quality responses. The combination of specifically designed prompts and the automatic iterative re-generation method performs the best, demonstrating that our method effectively enhances the quality of LLM responses, leading to improved performance in human activity recognition.

\subsection{\textbf{Parameter sensitivity}}
In our model, selecting the appropriate parameters for the sensor encoder is crucial, as it directly affects its suitability for deployment on mobile devices (e.g., ensuring the parameter size is sufficiently small). 
In this section, we analyze the impact of the embedding dimensions and the number of encoding layers on performance.

\subsubsection{\textbf{Effect of different number of attention heads}}
Fig.~\ref{fig:head} shows the performance comparison $w.r.t.$ the number of attention heads in the sensor encoder.
Overall, the performance is relatively stable under different numbers of attention heads.
We choose the number of attention heads of 2 because it provides good performance while reducing the model's parameter size.

\subsubsection{\textbf{Effect of different number of encoding layers}}
\begin{figure}[h]\centering
\begin{minipage}[h]{0.37\linewidth}
    \includegraphics[width=\linewidth, keepaspectratio=true]{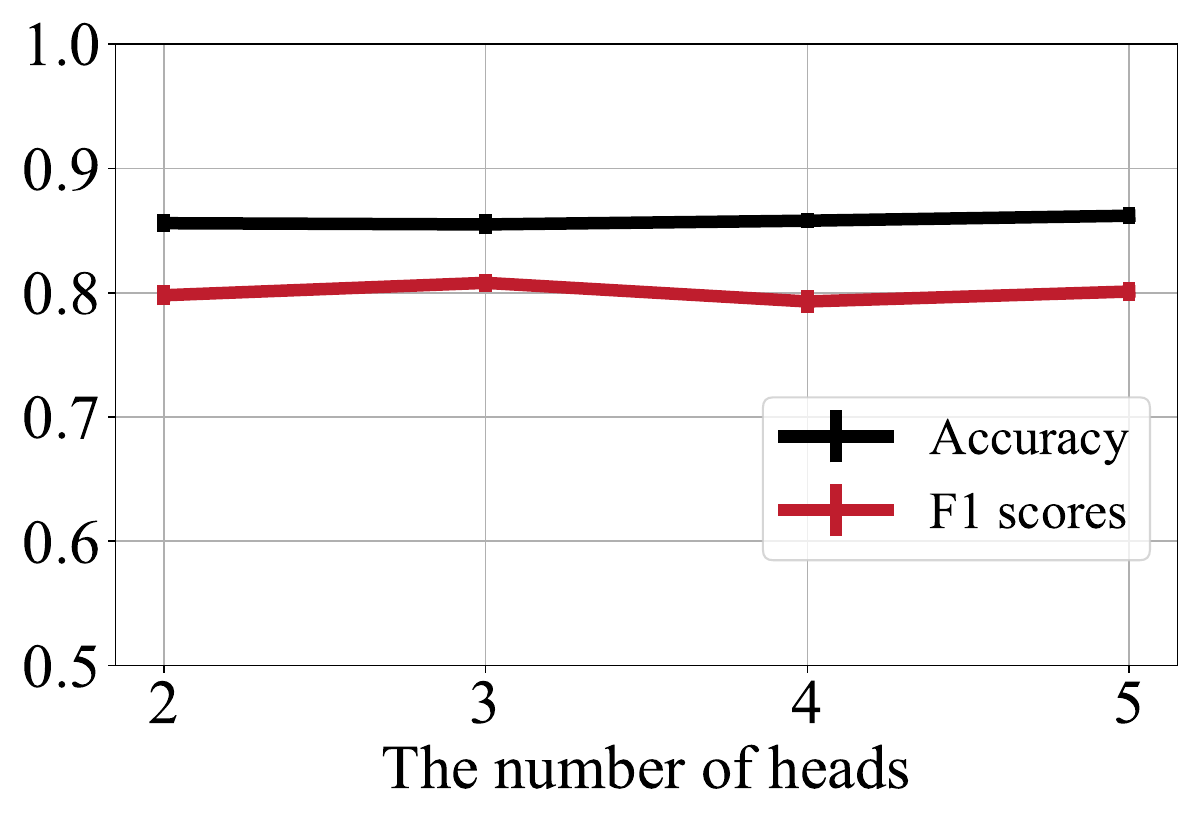}
    \vspace{-15pt}
\captionsetup{font={normalsize}}
    \caption{The effect of different number of attention heads}
    \label{fig:head}
\end{minipage}
\hspace{10pt}
\begin{minipage}[h]{0.37\linewidth}
    \includegraphics[width=\linewidth, keepaspectratio=true]{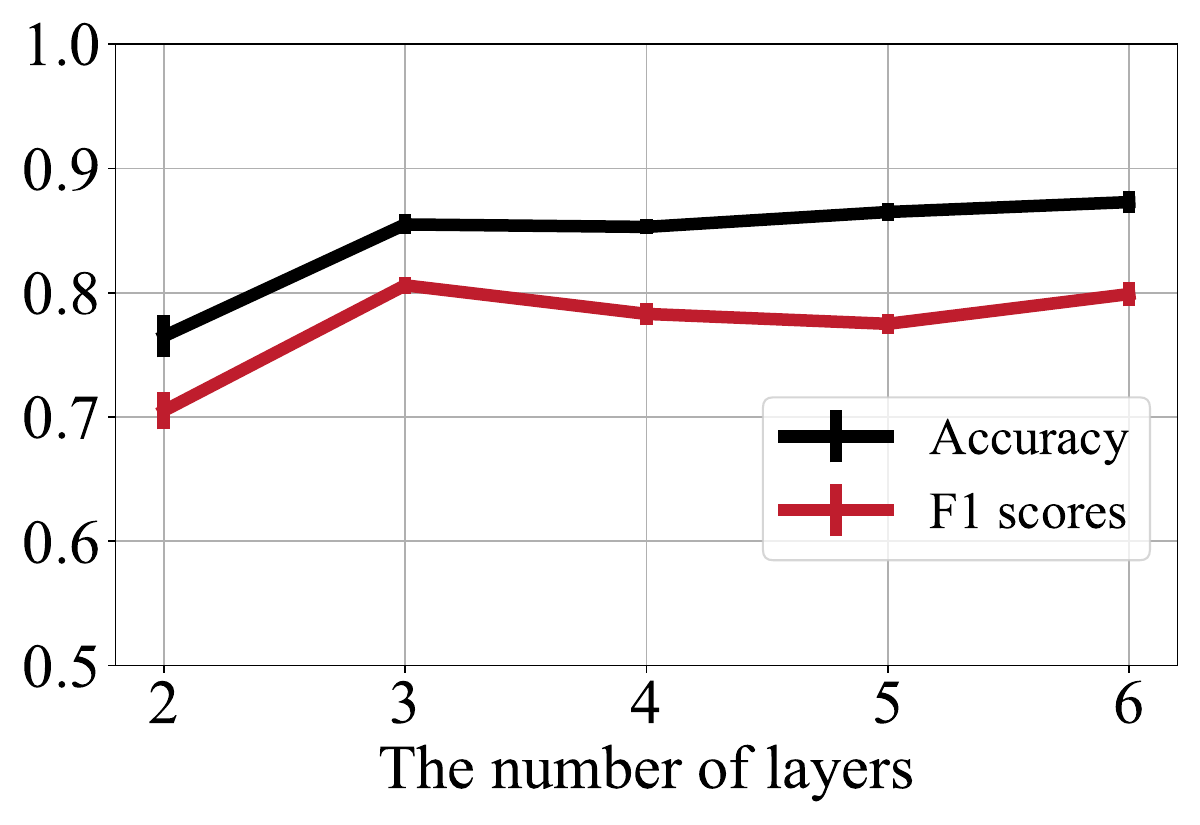}
    \vspace{-15pt}
  \captionsetup{font={normalsize}}
    \caption{The effect of different number of encoding layers}
    \label{fig:layer}
\end{minipage}
\vspace{-12pt}
\end{figure}

Fig.~\ref{fig:layer} shows the performance comparison with respect to different numbers of encoding layers in the sensor encoder. As observed, a smaller number of encoding layers (e.g., 2 layers) results in insufficient representation, whereas a larger number (e.g., 6 layers) increases model complexity. We select 3 encoding layers as they strike a balance, delivering good performance without significantly increasing model complexity.

\subsection{\textbf{Deployment analysis}}
To evaluate the suitability of the sensor encoder for mobile deployment, we analyze its parameter size and inference time. Deploying \name~on mobile devices involves three key steps: (i) developing an Android application, (ii) converting the model into a .pt file, and (iii) integrating it for processing accelerometer and gyroscope data for activity recognition.
For our case study, we select the Pixel 7 smartphone as the deployment platform. The sensor encoder, combined with the activity label embeddings, occupies approximately 40 MB of memory, of which 0.01 MB is allocated for the embeddings. To assess its real-time performance, we conduct 100 inference trials, measuring the time required to recognize an activity.
The results indicate an average processing time of 0.607 seconds, with the minimum and maximum times recorded at 0.716 seconds and 0.527 seconds, respectively.
For comparison, we evaluate the model’s inference efficiency on a single A5000 GPU with a batch size of 1. The average inference time for a single activity is 0.0007 seconds, with the minimum and maximum times recorded at 0.0006 seconds and 0.001 seconds, respectively.
Although inference on the smartphone is significantly slower than on the GPU, it remains within a practical range for real-world mobile applications, supporting real-time inference with an IMU sampling frequency of around 40 Hz (i.e., 0.025 seconds per reading).
This demonstrates that \name ~can operate effectively on mobile devices. 
To further enhance efficiency, future optimizations, such as quantization and hardware acceleration, could be explored.

\section{Discussion}
\noindent\textbf{Lessons learned:}
We summarize the following lessons to benefit future studies: 
(1) Introducing semantic interpretation can mitigate heterogeneity in cross-dataset HAR and offers the potential to recognize new activities, as indicated in Table~\ref{tb: overall performance} and Fig.~\ref{fig:New activity performance}.
(2) The success of LLMs in cross-dataset HAR is partly due to their ability to summarize patterns and represent them using language. This leads to a lower KL divergence between cross-dataset data distributions compared to raw data.
(3) Improving the quality of LLM responses through iterative re-generation, guided by the KL divergence, enhances HAR performance. The evaluation shows that generating high-quality responses has a direct positive impact on HAR accuracy.

\vspace{4pt}
\noindent\textbf{Comparison to other ways of utilizing LLMs:}
In addition to our method, there are two other potential methods for utilizing LLMs in HAR: training a foundation model specifically for HAR or fine-tuning an existing open-source LLM, such as Llama. 
A key advantage of our method lies in its low cost and flexibility.
On the one hand, our method does not require large amounts of data or computing resources for model training or fine-tuning. 
On the other hand, different LLMs can be flexibly integrated into our framework. As demonstrated in our evaluation, incorporating a high-quality LLM could further enhance the performance of our method.

\vspace{4pt}
\noindent\textbf{Other ways of improving LLMs responses:}
There are ongoing efforts in the AI community aimed at improving the trustworthiness of LLMs through methods such as retrieval-augmented generation and memory-augmented models. 
Our method offers a general framework for post-hoc factuality verification, where advancements in these AI methods can be seamlessly integrated into our framework.

\vspace{4pt}
\noindent\textbf{Limitations:}
(i) The process of obtaining semantic interpretations using LLMs is time-consuming, primarily due to the slow inference speed of current LLMs. We anticipate that future advancements in accelerating LLMs will enhance the efficiency of our method. 
(ii) Although we used four public datasets in the evaluation, they remain relatively small in scale with a limited number of activities. A larger dataset could provide a more comprehensive evaluation and further validate our method's effectiveness.
(iii) Although our evaluation on PAMAP2 simulates domain heterogeneity by splitting participants, it does not constitute a true cross-dataset validation. This design choice was motivated by the need to assess model performance on complex activities (e.g., vacuum cleaning, ironing), which are sparsely available and inconsistently defined across public datasets. Even though, we believe this setting is usable to evaluate the performance on complex activities when no additional datasets with the same complex activities are available.

\section{Conclusion}
In this work, we incorporate LLM-generated semantic interpretations of both sensor readings and activity labels to mitigate cross-dataset heterogeneity and to enable the recognition of unseen activities by mapping both sensor readings and activity labels into a shared semantic space. 
We design \name, a lightweight system that features a semantic interpretation generation process with an iterative re-generation mechanism to ensure output quality. Our two-stage training and inference framework allows the transfer of LLM capabilities to mobile and resource-constrained environments, enabling practical deployment scenarios.
We conduct comprehensive experiments across five public HAR datasets. The results demonstrate that our method consistently outperforms state-of-the-art approaches in cross-dataset activity recognition.
In future work, we plan to extend our framework to support open-set activity recognition, integrate physics-informed semantic interpretations and handle more complex, fine-grained activity patterns to further improve robustness and generalizability.

\bibliographystyle{ACM-Reference-Format}
\bibliography{mybibliography}
\newpage
\section*{Appendix}
\subsection*{Comparison to baselines with 5 simple activities}
Following the reviewer’s suggestion, we extend our cross-dataset experiments to include all relevant activities, treating sitting and standing as separate classes rather than merging them into a single still category. Specifically, the evaluated activities now include sitting, standing, walking, going upstairs, and going downstairs.
The new cross-dataset results for five activities are as shown in Table~\ref{tb: 5 activity performance}.

\begin{table*}[h]
\small
        \captionsetup{font={normalsize}}
\caption{Overall performance with 5 simple activities (walking, going upstairs, going downstairs,sitting and standing). Bolded scores indicate the best, while underlined scores represent the second best.}
\setlength\tabcolsep{1.7pt}
\renewcommand{\arraystretch}{1}
\label{tb: 5 activity performance}
\vspace{-10pt}
\begin{tabular}{c|c|cccccccccccc|c}
\toprule
\multirow{2}{*}{Method}      & Source & \multicolumn{3}{c|}{HHAR}                                       & \multicolumn{3}{c|}{UCI}                                         & \multicolumn{3}{c|}{Motion}                                   & \multicolumn{3}{c|}{Shoaib}                                  & \multirow{2}{*}{Average} \\ \cmidrule{2-14}
                             & Target & \multicolumn{1}{c}{UCI} & \multicolumn{1}{c}{Motion} & Shoaib & \multicolumn{1}{c}{HHAR} & \multicolumn{1}{c}{Motion} & Shoaib & \multicolumn{1}{c}{HHAR} & \multicolumn{1}{c}{UCI} & Shoaib & \multicolumn{1}{c}{HHAR} & \multicolumn{1}{c}{UCI} & Motion  & \\ \midrule
\multirow{2}{*}{DCNN}         & Accuracy    & \multicolumn{1}{c}{0.231}    & \multicolumn{1}{c}{0.154}       & 0.181     & \multicolumn{1}{c}{0.271}     & \multicolumn{1}{c}{0.175}       &  0.292      & \multicolumn{1}{c}{0.245}     & \multicolumn{1}{c}{0.342}    &  0.518      & \multicolumn{1}{c}{0.323}     & \multicolumn{1}{c}{0.462}    &  0.362       & 0.296 \\ 
                             & F1 scores    & \multicolumn{1}{c}{0.162}    & \multicolumn{1}{c}{0.091}       &  0.084     & \multicolumn{1}{c}{0.163}     & \multicolumn{1}{c}{0.107}       &  0.251      & \multicolumn{1}{c}{0.092}     & \multicolumn{1}{c}{0.246}    &   0.486     & \multicolumn{1}{c}{0.208}     & \multicolumn{1}{c}{0.405}    &  0.218     &  0.209\\ 
\multirow{2}{*}{Transformer} &  Accuracy      & \multicolumn{1}{c}{0.285}    & \multicolumn{1}{c}{0.266}       & 0.270       & \multicolumn{1}{c}{0.230}     & \multicolumn{1}{c}{0.244}       & 0.213       & \multicolumn{1}{c}{0.221}     & \multicolumn{1}{c}{0.246}    &   0.441     & \multicolumn{1}{c}{0.220}     & \multicolumn{1}{c}{0.372}    &   0.277     & 0.274 \\ 
                             & F1 scores       & \multicolumn{1}{c}{0.195}    & \multicolumn{1}{c}{0.198}       & 0.205       & \multicolumn{1}{c}{0.078}     & \multicolumn{1}{c}{0.158}       &   0.216     & \multicolumn{1}{c}{0.072}     & \multicolumn{1}{c}{0.182}    & 0.416       & \multicolumn{1}{c}{0.072}     & \multicolumn{1}{c}{0.287}    &  0.105       &  0.182\\ 
\multirow{2}{*}{LIMU-BERT} &  Accuracy      & \multicolumn{1}{c}{0.264}    & \multicolumn{1}{c}{0.214}       & 0.408       & \multicolumn{1}{c}{0.411}     & \multicolumn{1}{c}{0.230}       & 0.262      & \multicolumn{1}{c}{0.186}     & \multicolumn{1}{c}{0.129}    &   0.275     & \multicolumn{1}{c}{0.442}     & \multicolumn{1}{c}{0.313}    &   0.224     & 0.279 \\ 
                             & F1 scores       & \multicolumn{1}{c}{0.150}    & \multicolumn{1}{c}{0.139}       & 0.350       & \multicolumn{1}{c}{0.381}     & \multicolumn{1}{c}{0.075}       &   0.191     & \multicolumn{1}{c}{0.076}     & \multicolumn{1}{c}{0.064}    & 0.165       & \multicolumn{1}{c}{0.412}     & \multicolumn{1}{c}{0.234}    &  0.073      & 0.193 \\ \midrule
\multirow{2}{*}{UDAHAR}          & Accuracy       & \multicolumn{1}{c}{0.266}    & \multicolumn{1}{c}{0.316}       &  0.348     & \multicolumn{1}{c}{0.270}     & \multicolumn{1}{c}{0.173}       &  0.376      & \multicolumn{1}{c}{0.228}     & \multicolumn{1}{c}{0.286}    &   0.468     & \multicolumn{1}{c}{0.254}     & \multicolumn{1}{c}{0.496}    &   0.304     & 0.315 \\ 
                             &  F1 scores      & \multicolumn{1}{c}{0.128}    & \multicolumn{1}{c}{0.215}       &  0.250      & \multicolumn{1}{c}{0.125}     & \multicolumn{1}{c}{0.104}       &  0.339      & \multicolumn{1}{c}{0.074}     & \multicolumn{1}{c}{0.206}    &  0.447      & \multicolumn{1}{c}{0.137}     & \multicolumn{1}{c}{0.436}    &      0.155   & 0.218 \\ 
\multirow{2}{*}{SDMix}       & Accuracy       & \multicolumn{1}{c}{0.375}    & \multicolumn{1}{c}{0.349}       & 0.433      & \multicolumn{1}{c}{0.462}     & \multicolumn{1}{c}{0.390}       &   0.461     & \multicolumn{1}{c}{0.371}     & \multicolumn{1}{c}{0.327}    &   0.300     & \multicolumn{1}{c}{0.418}     & \multicolumn{1}{c}{0.197}    &  0.518       & 0.383 \\ 
                             & F1 scores       & \multicolumn{1}{c}{0.244}    & \multicolumn{1}{c}{0.273}       & 0.370      & \multicolumn{1}{c}{0.395}     & \multicolumn{1}{c}{0.245}       &  0.441      & \multicolumn{1}{c}{0.300}     & \multicolumn{1}{c}{0.281}    &    0.172    & \multicolumn{1}{c}{0.386}     & \multicolumn{1}{c}{0.135}    &  0.435          & 0.306 \\ 
\multirow{2}{*}{UniMTS}       & Accuracy       & \multicolumn{1}{c}{0.439}    & \multicolumn{1}{c}{0.457}       & 0.482      & \multicolumn{1}{c}{0.441
}     & \multicolumn{1}{c}{0.402}       &   0.511     & \multicolumn{1}{c}{0.407}     & \multicolumn{1}{c}{0.437}    &   0.514     & \multicolumn{1}{c}{0.492}     & \multicolumn{1}{c}{0.402}    &  0.553      & 0.461 \\ 
                             & F1 scores       & \multicolumn{1}{c}{0.355}    & \multicolumn{1}{c}{0.361}       & 0.383       & \multicolumn{1}{c}{0.392}     & \multicolumn{1}{c}{0.328}       &  0.405      & \multicolumn{1}{c}{0.301}     & \multicolumn{1}{c}{0.382}    &    0.416    & \multicolumn{1}{c}{0.354}     & \multicolumn{1}{c}{0.327}    &  0.468         & 0.372  \\ 
\multirow{2}{*}{UniHAR}      &  Accuracy    & \multicolumn{1}{c}{\underline{0.575}}    & \multicolumn{1}{c}{\underline{0.523}}       & \underline{0.489}       & \multicolumn{1}{c}{\underline{0.475}}     & \multicolumn{1}{c}{\underline{0.569}}       &  \underline{0.628}      & \multicolumn{1}{c}{\underline{0.553}}     &  \multicolumn{1}{c}{\underline{0.570}}    &  \underline{0.551}     & \multicolumn{1}{c}{\underline{0.464}}     & \multicolumn{1}{c}{\underline{0.569}}    & \underline{0.579}   & \underline{0.545} \\ 
                             & F1 scores       & \multicolumn{1}{c}{\underline{0.470}}    & \multicolumn{1}{c}{\underline{0.399}}       & \underline{0.413}       & \multicolumn{1}{c}{\underline{0.421}}     & \multicolumn{1}{c}{\underline{0.489}}       &  \underline{0.586}      & \multicolumn{1}{c}{\underline{0.457}}     & \multicolumn{1}{c}{\underline{0.510}}    & \underline{0.480}      & \multicolumn{1}{c}{\underline{0.397}}     & \multicolumn{1}{c}{\underline{0.491}}    & \underline{0.501}    &  \underline{0.467}\\ \midrule
\multirow{2}{*}{\name{}}        & Accuracy       & \multicolumn{1}{c}{\textbf{0.629}}    & \multicolumn{1}{c}{\textbf{0.585}}       &  \textbf{0.557}      & \multicolumn{1}{c}{\textbf{0.576}}     & \multicolumn{1}{c}{\textbf{0.624}}       &  \textbf{0.631}      & \multicolumn{1}{c}{\textbf{0.587}}     & \multicolumn{1}{c}{\textbf{0.611}}    &  \textbf{0.558}     & \multicolumn{1}{c}{\textbf{0.561}}     & \multicolumn{1}{c}{\textbf{0.596}}    &  \textbf{0.580}    &  \textbf{0.591} \\ 
                             &  F1 scores     & \multicolumn{1}{c}{\textbf{0.510}}    & \multicolumn{1}{c}{\textbf{0.461}}       &  \textbf{0.447}      & \multicolumn{1}{c}{\textbf{0.443}}     & \multicolumn{1}{c}{\textbf{0.528}}       &   \textbf{0.588}    & \multicolumn{1}{c}{\textbf{0.502}}    & \multicolumn{1}{c}{\textbf{0.527}}    &   \textbf{0.484}   & \multicolumn{1}{c}{\textbf{0.492}}     & \multicolumn{1}{c}{\textbf{0.512}}    &     \textbf{0.505}    & \textbf{0.497}  \\ \bottomrule
\end{tabular}
\end{table*}
\end{document}